\documentclass{article} 
\usepackage{iclr2027_conference,times}

\usepackage{amsmath,amsfonts,bm}

\def\eqref#1{equation~\ref{#1}}

\def\1{\bm{1}}

\def\rx{{\textnormal{x}}}

\def\vtheta{{\bm{\theta}}}

\def\vh{{\bm{h}}}

\def\vv{{\bm{v}}}

\def\mM{{\bm{M}}}

\DeclareMathAlphabet{\mathsfit}{\encodingdefault}{\sfdefault}{m}{sl}
\SetMathAlphabet{\mathsfit}{bold}{\encodingdefault}{\sfdefault}{bx}{n}

\usepackage[utf8]{inputenc} 
\usepackage[T1]{fontenc}    
\usepackage{hyperref}       
\usepackage{url}            
\usepackage{booktabs}       
\usepackage{amsfonts}       
\usepackage{nicefrac}       
\usepackage{microtype}      
\usepackage{xcolor}         
\usepackage{graphicx}
\usepackage{colortbl}
\usepackage{booktabs}
\usepackage{subfig}
\usepackage{lscape}   
\usepackage{amsmath}
\usepackage{listings}
\usepackage{xcolor}
\usepackage{tabularx}
\usepackage{caption}
\usepackage{amssymb}
\usepackage{subcaption}
\definecolor{gain}{HTML}{2E7D32}

\definecolor{codegreen}{rgb}{0,0.6,0}
\definecolor{codegray}{rgb}{0.5,0.5,0.5}
\definecolor{codepurple}{rgb}{0.58,0,0.82}
\definecolor{backcolour}{rgb}{0.95,0.95,0.92}

\lstdefinestyle{mystyle}{
    backgroundcolor=\color{backcolour},   
    commentstyle=\color{codegreen},
    keywordstyle=\color{magenta},
    numberstyle=\tiny\color{codegray},
    stringstyle=\color{codepurple},
    basicstyle=\ttfamily\footnotesize,
    breakatwhitespace=false,         
    breaklines=true,                 
    captionpos=b,                    
    keepspaces=true,                 
    numbers=left,                    
    numbersep=5pt,                  
    showspaces=false,                
    showstringspaces=false,
    showtabs=false,                  
    tabsize=2
}

\title{LightMTP:\\Lightweight Latent Multi-Token Prediction}

\author{\normalfont
\begin{tabular*}{\dimexpr\textwidth-2\tabcolsep\relax}[t]{@{}l@{\extracolsep{\fill}}l@{}}
\bf Tamara Czinczoll & \bf Julie Kallini\\
Hasso Plattner Institute / University of Potsdam & Stanford University\\
Potsdam, Germany & Stanford, USA\\
\texttt{tamara.czinczoll@hpi.de} & \texttt{kallini@stanford.edu}\\
\rule{0pt}{24pt}\bf Gerard de Melo & \bf Chen Shani\\
Hasso Plattner Institute / University of Potsdam & Tel-Aviv University\\
Potsdam, Germany & Tel Aviv, Israel\\
\texttt{gerard.demelo@hpi.de} & \texttt{cshani@tauex.tau.ac.il}%
\end{tabular*}}

\newcommand{\pkey}[1]{\textcolor[HTML]{#1}{$\blacksquare$}\,}

\iclrfinalcopy 
\begin{document}

\maketitle
\begin{abstract}
Next-token prediction (NTP) is the standard pretraining objective for large language models, yet it provides an explicit training signal only for the \emph{immediate} next token, which can lead models to exploit local patterns instead of capturing longer-range structure and ideas. Multi-token prediction (MTP) addresses this by training models to predict several future tokens. However, existing MTP methods often introduce a large number of new parameters with limited improvements in downstream performance. Latent MTP approaches address this efficiency issue by encoding future tokens into a vector representation. However, these approaches usually rely on external helper models for future token encoding. We propose LightMTP, a lightweight, i.e., parameter-efficient, latent MTP approach that bootstraps the future token representations from the model's \emph{own} hidden states. Our two LightMTP variants extend supervision to more future tokens without requiring the additional computational overhead of conventional MTP nor the external supervision latent MTP normally relies on. LightMTP adds at most 1\% extra parameters, retains better performance on general language modeling benchmarks, and achieves similar gains in planning, coding, and reasoning.
\end{abstract}

\begin{figure}[h]
    \centering
    \includegraphics[width=\linewidth]{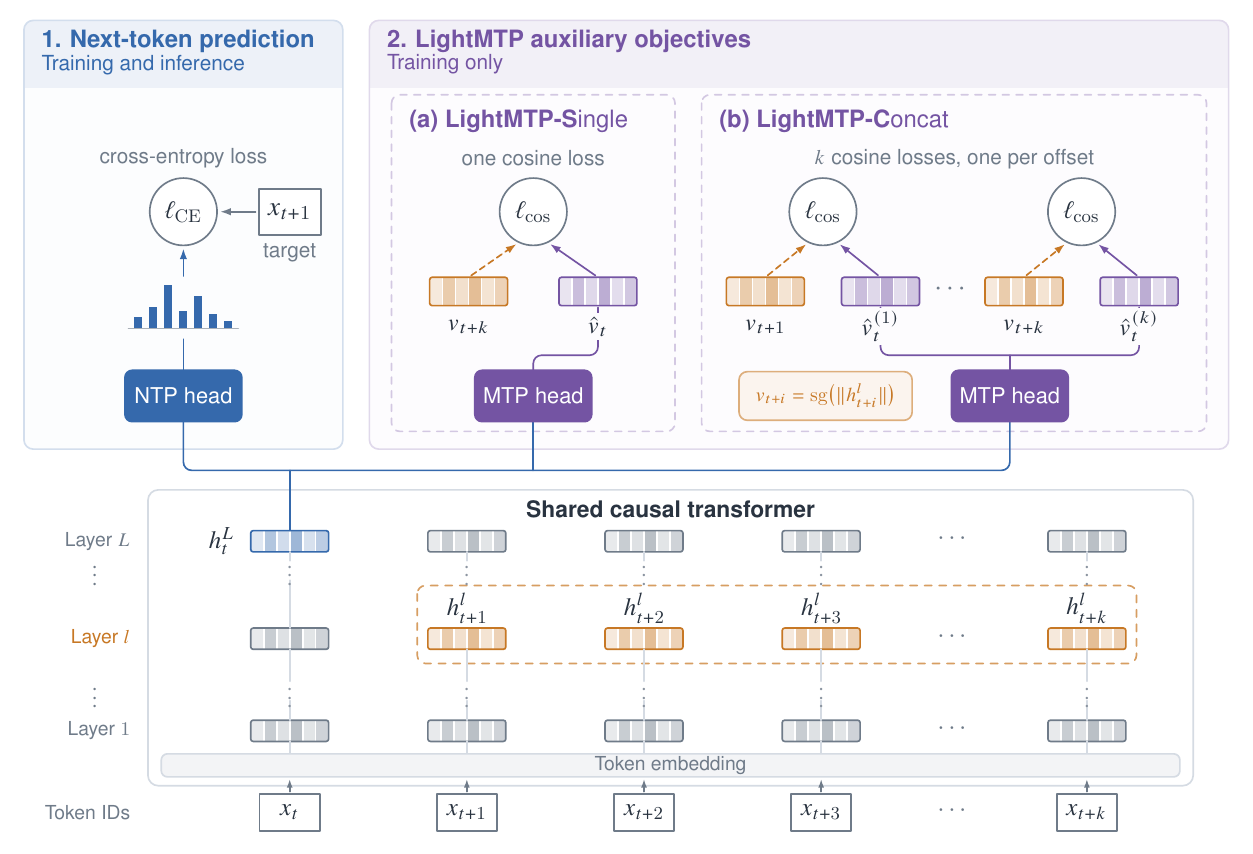}
    \caption{LightMTP passes the input through a shared transformer trunk and uses the last hidden state at each position to 1) perform next-token prediction via a linear NTP head, and 2) predict a latent representation of the next $k$ target vectors stemming from the model's own future hidden states via an MTP head. a) In our approach's single-target variant, the model predicts the single hidden state vector in $k$ steps. b) In the version with concatenated targets, each intermediate vector is additionally used to supervise the model.}
    
    \label{fig:overview-mtp}
\end{figure}

\section{Introduction}

Next-token prediction (NTP) is the standard pretraining objective for large language models. It requires no supervision beyond raw text, computes the loss at every position in a single teacher-forced forward pass, and improves predictably with scale \citep{kaplan_scaling_2020, hoffmann_empirical_2022}. However, NTP provides an explicit training signal only for the immediate next token. Information about more distant tokens is learned only insofar as it helps predict the next one, which can lead models to exploit local patterns instead of capturing longer-range structure. \citet{bachmann_pitfalls_2024} show that teacher-forced NTP can fail to learn even simple planning tasks.

Multi-token prediction (MTP) addresses this limitation by additionally training the model to predict the tokens $x_{t+2}, \dots, x_{t+k}$ from its representation at position $t$. This densifies the training signal and rewards representations that encode information beyond the next token \citep{gloeckle_better_2024, deepseek-ai_deepseek-v3_2025}. Most MTP methods operate at the token level, attaching one prediction module per future offset that outputs a vocabulary distribution. However, this is costly: each offset adds a vocabulary projection, so compute and memory grow with both $k$ and the vocabulary size, and many designs add highly parametrized transformer layers per offset. Additionally, token-level MTP can also interfere with next-token prediction. \citet{gloeckle_better_2024} find that it hurts models below roughly 3B parameters.

Latent MTP methods avoid the vocabulary projection by predicting dense representations of future tokens. They typically obtain their targets from auxiliary encoders or separately trained helper models \citep{yin_semformer_2024, mahajan_beyond_2026}. We propose \textbf{LightMTP, a lightweight latent MTP objective that bootstraps its targets from the model's own hidden states} (Figure~\ref{fig:overview-mtp}). A small MLP head reads the last hidden state at position $t$ and directly predicts the intermediate layer states of future positions up to $k$ steps ahead. Because the targets are latent, the model is encouraged to encode information about upcoming tokens without committing to specific ones. LightMTP requires no auxiliary model, avoids projecting onto the vocabulary, and adds at most 1\% parameters. The head is discarded after training, so inference remains unchanged.

We compare LightMTP with NTP, three token-level MTP methods, and a latent MTP baseline at model sizes up to 1.5B parameters. LightMTP produces hidden states that are substantially more predictive of future tokens than NTP, while incurring a smaller next-token performance cost than the MTP baselines. On Countdown, a planning task, LightMTP improves over NTP by up to 10 points and more often preserves target reachability. On code generation and math reasoning, it follows output formats more consistently and makes fewer structural errors than NTP. Unlike the token-level MTP baselines, LightMTP also retains NTP-level performance on general language modeling benchmarks, supporting its use as a general-purpose pretraining objective.

\section{Related Work}
\paragraph{NTP vs MTP.}
Predicting several future tokens predates decoder-only LLMs, having been explored both to accelerate decoding \citep{stern_blockwise_2018} and as a pretraining objective in ProphetNet \citep{qi_prophetnet_2020} to reduce local overfitting.
\citet{gloeckle_better_2024} first applied MTP to decoder-only transformer models by assigning each future token position its own head, with all positions sharing the unembedding matrix and transformer's trunk. For simplicity, their method does not fully model the joint probability of the next $k$ tokens but approximates it by assuming conditional independence given the previous context. This led \citet{basharin_faster_2025} to propose using a rank-$r$ probability decomposition to better approximate the joint future token distribution, subsequently achieving lower joint token loss on one dataset. However, they do not demonstrate the impact on more general pretraining benchmarks. The first to pretrain a large-scale frontier model with MTP, DeepSeek-v3 models the joint token probability via sequential per-position modules that partially share parameters \citep{deepseek-ai_deepseek-v3_2025}. However, they only train with one additional token ($k=2$). While it has been shown that LLMs plan internally for some future tokens beyond the immediate next one \citep{pal_future_2023}, this effect is limited and might benefit from more explicit supervision during training.

\paragraph{Latent MTP.}
Latent MTP methods differ mainly in where their targets come from. Semformer \citep{yin_semformer_2024} trains planning tokens to predict an autoencoder's representation of the response, next-concept prediction \citep{liu_next_2026} predicts vector-quantized concepts spanning several tokens, and Future Summaries \citep{mahajan_beyond_2026} predicts the hidden states of a separately pretrained right-to-left language model. These targets require auxiliary encoders, codebooks, or helper models, which add parameters and training stages. Similar to our work, NextLat \citep{teoh_next-latent_2026} also regress the model's own stop-gradient hidden states. They train a transition model that must predict each next hidden state from only the current state and the next token to obtain more compact, less shortcut-prone representations that generalize and plan better. LightMTP differs in three ways. It predicts states up to $k$ steps ahead from the current state alone, without access to the intervening ground-truth tokens, learning to anticipate rather than model transitions. Furthermore, it predicts the future in a single parallel head call rather than a sequential rollout. Lastly, it takes its targets from an intermediate layer and uses only a regression loss, avoiding any projection onto the vocabulary.

\section{LightMTP: Lightweight Latent Multi-Token Prediction}
We now describe the architecture of LightMTP. 

\subsection{Latent Prediction Head and Training Objective}
\label{sec:lightmtp}
A transformer trunk with $L$ layers maps a prefix $\rx_{\leq t}$ to hidden states $\vh_t^{l}$ for $l=1,\dots,L$. Standard NTP trains a linear head $\mM\in\mathbb{R}^{|V|\times d}$ on $\vh_t^{L}$ with a cross-entropy loss $\mathcal{L}_{\text{NTP}}$. As visualized in Figure~\ref{fig:overview-mtp}, we add a second head $g_{\bm{\phi}}$, parametrized separately from the trunk parameters $\vtheta$, which reads $\vh_t^{L}$ and predicts hidden states the trunk itself produces at future positions. During training, this head shapes the transformer
trunk’s parameters that are shared with the NTP objective, leading to hidden states that are more informed about future tokens. In our experiments, $g_{\bm{\phi}}$ is a two-layer MLP with a GELU nonlinearity.

\textbf{Targets.} As target vectors, we consider the trunk's own hidden states $\vh_\tau^{l}$ at a chosen target layer $0\leq l\leq L$, taken at some future position $\tau>t$. As these are produced by the same trunk that is being trained, we stop gradients from flowing into $\vtheta$ through the target branch, denoted $\mathrm{stopgrad}(\vh_\tau^{l})$; the head's own prediction error still backpropagates into $\vtheta$ through $\vh_t^{L}$, the input to $g_{\bm{\phi}}$.

\textbf{Loss.} We normalize both the predicted and the target vectors. Given a normalized predicted vector $\hat \vv\in\mathbb{R}^d$ and a normalized, stop-gradient target $\vv\in\mathbb{R}^d$, we consider a cosine loss function $\ell(\hat \vv, \vv)$:
\begin{align}
\ell_{\cos}(\hat \vv, \vv) &= 1 - \frac{\hat \vv \cdot \vv}{\lVert \hat \vv \rVert_2\, \lVert \vv \rVert_2}
\end{align}

\textbf{Encouraging representational diversity in the target layer.} Because the target layer's representations are themselves shaped by training, the model could potentially make all positions similar, trivializing the MTP loss without transferring useful information. While the NTP objective encourages some representational diversity, we additionally experiment with \emph{Variance Regularization}, a within-sequence variant of cosine regularization \citep{gao_representation_2019} that penalizes cross-positional similarity of the normalized target-layer representations, computed per batch element $b\in\{1,\dots,B\}$ over the sequence of length $T$:
\begin{align}
\hat{\vh}_{b,t}^l &= \mathrm{norm}(\vh_{b,t}^{l}),  \ \ \ \ \ \ \ \bar{\vh}_b^l = \frac{1}{T}\sum_{t=1}^{T} \hat{\vh}_{b,t}^l, \ \ \ \ \ \ \
\mathcal{L}_{\text{VarReg}} = \frac{1}{B}\sum_{b=1}^{B} \lVert \bar{\vh}_b^l \rVert_2
\end{align}
A large mean-vector norm $\lVert \bar \vh_b^l \rVert_2$ indicates that positions share a common direction rather than being spread out, so penalizing it encourages position-specific representations.

\textbf{Overall loss.} Both variants below are trained with
\begin{align}
    \mathcal{L}_{\text{total}} = \mathcal{L}_{\text{NTP}} + \lambda_{\text{MTP}} \,\mathcal{L}_{\text{MTP}} + \lambda_{\text{VarReg}}\, \mathcal{L}_{\text{VarReg}},
\end{align}
where $\mathcal{L}_{\text{NTP}}$ is standard Cross-Entropy Loss and $\mathcal{L}_{\text{MTP}}$ is defined separately for each variant.

\subsection{LightMTP-S: Single-Target Vector Prediction}
\label{sec:single-target}
The simplest instantiation of $g_{\bm{\phi}}$ predicts a single future representation $k$ steps ahead, rather than modeling every intermediate position. Concretely, $g_{\bm{\phi}}: \mathbb{R}^d \to \mathbb{R}^d$ maps the last hidden state at position $t$ to a prediction of the target-layer hidden state $k$ steps in the future:
\begin{align}
\hat \vv_{t} = \mathrm{norm}\big(g_{\bm{\phi}}(\vh_t^{L})\big), \qquad \vv_{t+k} = \mathrm{norm}\big(\mathrm{stopgrad}(\vh_{t+k}^{l})\big).
\end{align}
The MTP loss averages the per-position loss over the sequence:
\begin{align}
\mathcal{L}_{\text{MTP}}^{\text{single}} = \frac{1}{T-k}\sum_{t=1}^{T-k} \ell(\hat \vv_t, \vv_{t+k}).
\end{align}
Note that $k$ here only sets the lookahead \emph{offset} of a single target vector. It does not aggregate information from the intervening $k-1$ positions but relies on the vector's contextualized representation for intermediate information.

\subsection{LightMTP-C: Concatenated Multi-Target Vector Prediction}
\label{sec:concat-target}
Rather than committing to a single future point, we can instead task $g_{\bm{\phi}}$ to \emph{jointly} predict the hidden states for all $k$ future positions by concatenating them along the channel dimension. Here $g_{\bm{\phi}}:\mathbb{R}^d \to \mathbb{R}^{k\cdot d}$ produces a single $k\cdot d$-dimensional output, which we split into $k$ length-$d$ slices $\hat \vv_t^{(1)}, \dots, \hat \vv_t^{(k)}$, each compared against its corresponding future hidden state:
\begin{align}
[\hat \vv_t^{(1)}; \dots ; \hat \vv_t^{(k)}] &= g_{\bm{\phi}}(\vh_t^{L}) \in \mathbb{R}^{k\cdot d}, \\
\hat \vv_t^{(i)} \leftarrow \mathrm{norm}\big(\hat \vv_t^{(i)}\big), \qquad \vv_{t+i} &= \mathrm{norm}\big(\mathrm{sg}(\vh_{t+i}^{l})\big), \qquad i=1,\dots,k.
\end{align}
Each slice is normalized independently, so the loss scale does not depend on $k$. We compute one loss term per future step $i$, averaged over positions $t$ that are valid for that step, i.e., where $t+i$ remains within the same packed document and before the sequence ends ($m_{t,i}=1$; this lets end-of-sequence positions still train on whichever of their $k$ future steps are in range, instead of discarding them entirely):
\begin{align}
\ell_i = \frac{\sum_t m_{t,i}\, \ell\big(\hat \vv_t^{(i)}, \vv_{t+i}\big)}{\sum_t m_{t,i}}.
\end{align}
The $k$ per-step losses are then aggregated, using uniform weights, as
\begin{align}
\mathcal{L}_{\text{MTP}}^{\text{concat}} = \sum_{i=1}^k \ell_i,
\end{align}
targeting the same $k$ future positions as $k$ applications of the single-target head at offsets $1,\dots,k$.

\section{Experimental Setup}
We investigate the impact of latent MTP via four avenues: 1) models' hidden states, 2) general language modeling performance, 3) models' local planning abilities and, 4) structural tasks via coding and reasoning.

\paragraph{Models.}We pretrain 24-layer decoder-only models with 1.4–1.5B parameters (Table~\ref{tab:param_count_complete}), a scale at which MTP has often failed to help. We compare five baselines, namely NTP, two MTP variants in the style of \cite{gloeckle_better_2024} from Meta\footnote{We denote ``Meta'' as the paper's reported implementation with an extra Transformer layer per MTP head, and ``Meta-L'' as their linear variant described in the paper's appendix.}, DeepSeek-style MTP, and Future Summaries (see §\ref{app:baseline-setup} for details) with the two LightMTP variants, LightMTP-S(ingle) and LightMTP-C(oncat). Unless stated otherwise, LightMTP uses target layer $l=16$ (two-thirds of depth; we use the same ratio at depths 12 and 18) and horizon $k=4$, following related work. We also report $k=10$ in Table~\ref{tab:d24_headline}, and a full horizon sweep in Appendix~\ref{sec:horizon-k}. To isolate the effect of the variance regularizer, we train every LightMTP variant with and without VarReg ($\lambda_{\text{VarReg}}=0.1$). In the probing analysis (§\ref{sec:rep_future_tok}), we also include target layer $l=2$ to test whether VarReg's effect on anticipation holds across target layers. We do not use $l=2$ for downstream evaluation.
Since NTP is the essential baseline, it is trained over five seeds, all other models over three.

\paragraph{Pretraining.}
We pretrain models with a parameter-dependent token budget. (Results for models with 12 and 18 layers, respectively, are reported in Appendices~\ref{app:d12} and \ref{app:d18}.) This ensures that Meta's and DeepSeek's approaches, which add a substantial amount of MTP parameters, are not undertrained compared to the other models, but it leads to them seeing +20.7\%/+11.6\% (Meta-L/Meta) and +13.6\% (DeepSeek) more tokens, respectively. We use a global batch size of $524{,}288$ tokens and a tokenizer of size $32{,}768$. For an overview of the model's architecture, used hyperparameters, and parameter and token budgets, see Appendix~\ref{app:setup}.

\section{Results}
We organize the results around the cost of better future token representation (\S~\ref{sec:rep_future_tok}), general language modeling benchmarks (\S~\ref{sec:bench}), planning (\S~\ref{sec:planning}), and reasoning and coding (\S~\ref{sec:structure}) abilities.

\begin{figure}[t]
\centering
\includegraphics[width=\textwidth]{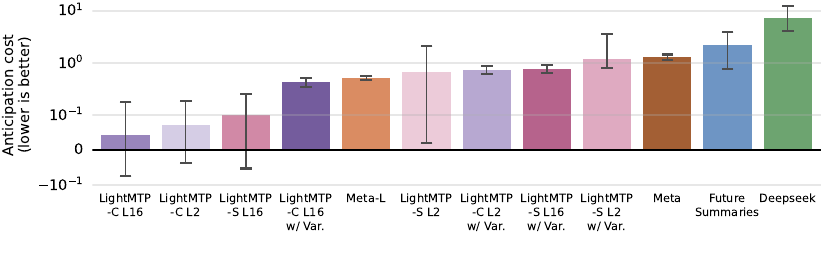}
\caption{\textbf{LightMTP pays fewer bits-per-byte --} Plot of anticipation cost, which quantifies how many bits-per-byte the models pay per total future token prediction improvement up to position $k$  for $k{\leq}10$ (lower is better). The top four most cost-efficient models are LightMTP variants.\label{fig:anticipation-cost}}
\label{fig:anticipation-cost}
\end{figure}

\subsection{Future Token Anticipation} \label{sec:rep_future_tok}

LightMTP, as well as the other MTP approaches, adds an explicit local lookahead pressure to the pretraining objective. To measure how this additional objective affects the models' representations, we train linear probes to predict the token at $t+k$ using the models' frozen last hidden-layer outputs on a held-out FineWeb-Edu shard. Since the last hidden layer directly receives gradients from the MTP heads, we expect MTP models to surface more future-token information in this layer, making future tokens more easily linearly decodable. The resulting probing results, shown in Appendix~\ref{app:probing}, reveal a tradeoff: \textbf{MTP methods generally improve future-token prediction ($k\geq2$), but this comes at a cost to NTP ($k=1$)}, most prominently for Meta's and DeepSeek's MTP approaches. Variance-based regularization further improves future-token anticipation for LightMTP, but also increases this cost.

To quantify this tradeoff, we introduce \emph{anticipation cost}, defined as the degradation in NTP divided by the gain in future-token prediction:
$$
\text{Anticipation Cost}
=
\frac{\Delta\mathrm{Bpb}_{k=1}}
{\Delta\mathrm{Acc}_{k>1}}.
$$
Here, $\Delta\mathrm{Bpb}_{k=1}$ is the increase in next-token bits-per-byte relative to NTP, while $\Delta\mathrm{Acc}_{k>1}$ is the corresponding gain in future-token prediction accuracy. Thus, anticipation cost captures how much next-token performance a method gives up for each unit of additional future-token information encoded in its representations. Figure~\ref{fig:anticipation-cost} visualizes anticipation cost across MTP methods. Four LightMTP variants (LightMTP-C L16 and L2, LightMTP-S L16, and LightMTP-C L16 with variance regularization) exhibit the lowest anticipation cost of all methods, with Meta-L in fifth place. Future Summaries and DeepSeek's approach have anticipation costs two orders of magnitude higher than the best LightMTP variant. Overall, these results suggest that, at this model scale, increasing future-token anticipation eventually comes at the cost of NTP quality, and that this tradeoff may be a key driver of the general language modeling impairment observed for baseline MTP methods in Section~\ref{sec:bench}. Notably, \textbf{LightMTP achieves substantially more future-token prediction with much smaller NTP degradation}, suggesting that its intermediate-layer supervision provides a more favorable tradeoff between anticipation and next-token modeling.

\subsection{Language Modeling Performance: LightMTP Shows Least Degradation}\label{sec:bench}
Commonly used to evaluate models' reasoning, world knowledge, and long-range capabilities, we report performance on the CORE task suite \citep{li_datacomp-lm_2025} and LAMBADA\footnote{LAMBADA is also included in the CORE suite.} \citep{paperno_lambada_2016}. CORE comprises 22 datasets spanning a broad range of language model capabilities. 
We include LAMBADA, since it tests long-range context use through a cloze task, in which the final word of a short story can only be predicted by incorporating context beyond the final sentence.

\begin{table}[t]
\centering
\setlength{\tabcolsep}{2pt}
\resizebox{\ifdim\width>\textwidth\textwidth\else\width\fi}{!}{
\begin{tabular}{lcccc}
\toprule
\textbf{Method} & \begin{tabular}[b]{@{}c@{}}\textbf{FineWeb-Edu}\\\textbf{(PPL)}\end{tabular} & \begin{tabular}[b]{@{}c@{}}\textbf{Lambada}\\\textbf{(PPL)}\end{tabular} & \begin{tabular}[b]{@{}c@{}}\textbf{Lambada}\\\textbf{(Acc)}\end{tabular} & \begin{tabular}[b]{@{}c@{}}\textbf{CORE-avg}\end{tabular} \\
\midrule
NTP & \textbf{1.683}$_{\pm0.001}$ & 16.00$_{\pm0.72}$ & 43.59$_{\pm0.85}$ & \underline{40.43}$_{\pm0.68}$ \\
Meta & \cellcolor[rgb]{0.947,0.802,0.802} 1.712$_{\pm0.003}$ & \cellcolor[rgb]{0.975,0.906,0.906} 18.66$_{\pm0.84}$ & \cellcolor[rgb]{0.973,0.899,0.899} 42.14$_{\pm0.91}$ & \cellcolor[rgb]{0.965,0.867,0.867} 39.31$_{\pm0.69}$ \\
Meta-L & \cellcolor[rgb]{0.972,0.894,0.894} 1.698$_{\pm0.001}$ & \cellcolor[rgb]{0.980,0.925,0.925} 18.13$_{\pm0.48}$ & \cellcolor[rgb]{0.986,0.948,0.948} 42.84$_{\pm0.70}$ & \cellcolor[rgb]{0.983,0.936,0.936} 39.89$_{\pm0.42}$ \\
DeepSeek & \cellcolor[rgb]{0.855,0.452,0.452} 1.762$_{\pm0.038}$ & \cellcolor[rgb]{0.855,0.452,0.452} 31.49$_{\pm11.42}$ & \cellcolor[rgb]{0.855,0.452,0.452} 35.68$_{\pm4.28}$ & \cellcolor[rgb]{0.855,0.452,0.452} 35.81$_{\pm2.60}$ \\
Future Summaries & \cellcolor[rgb]{0.995,0.981,0.981} 1.685$_{\pm0.002}$ & \cellcolor[rgb]{0.467,0.669,0.477} \textbf{15.57}$_{\pm0.33}$ & \cellcolor[rgb]{0.467,0.669,0.477} \textbf{44.43}$_{\pm0.45}$ & \cellcolor[rgb]{0.995,0.979,0.979} 40.25$_{\pm0.37}$ \\
\midrule
LightMTP-C & \cellcolor[rgb]{0.999,0.997,0.997} \textbf{1.683}$_{\pm0.000}$ & \cellcolor[rgb]{0.997,0.988,0.988} 16.35$_{\pm0.20}$ & \cellcolor[rgb]{0.993,0.974,0.974} 43.22$_{\pm0.42}$ & \cellcolor[rgb]{0.996,0.985,0.985} 40.30$_{\pm0.23}$ \\
\hspace{1em}+VarReg & \cellcolor[rgb]{0.988,0.954,0.954} 1.689$_{\pm0.000}$ & \cellcolor[rgb]{0.992,0.970,0.970} 16.84$_{\pm0.41}$ & \cellcolor[rgb]{0.983,0.935,0.935} 42.65$_{\pm0.51}$ & \cellcolor[rgb]{0.992,0.971,0.971} 40.18$_{\pm0.43}$ \\
\hspace{1em}k=10 & \cellcolor[rgb]{0.995,0.982,0.982} 1.685$_{\pm0.005}$ & \cellcolor[rgb]{0.955,0.972,0.956} 15.97$_{\pm1.04}$ & \cellcolor[rgb]{0.703,0.815,0.709} \underline{44.06}$_{\pm1.47}$ & \cellcolor[rgb]{0.467,0.669,0.477} \textbf{40.57}$_{\pm0.34}$ \\
\cmidrule{1-5}
LightMTP-S & \cellcolor[rgb]{0.999,0.995,0.995} \underline{1.684}$_{\pm0.000}$ & \cellcolor[rgb]{0.997,0.990,0.990} 16.28$_{\pm0.48}$ & \cellcolor[rgb]{0.995,0.981,0.981} 43.31$_{\pm0.71}$ & \cellcolor[rgb]{0.991,0.964,0.964} 40.13$_{\pm0.67}$ \\
\hspace{1em}+VarReg & \cellcolor[rgb]{0.979,0.923,0.923} 1.694$_{\pm0.002}$ & \cellcolor[rgb]{0.986,0.946,0.946} 17.53$_{\pm0.35}$ & \cellcolor[rgb]{0.977,0.913,0.913} 42.33$_{\pm0.14}$ & \cellcolor[rgb]{0.996,0.984,0.984} 40.29$_{\pm0.46}$ \\
\hspace{1em}k=10 & \cellcolor[rgb]{0.996,0.984,0.984} 1.685$_{\pm0.001}$ & \cellcolor[rgb]{0.896,0.936,0.898} \underline{15.92}$_{\pm0.46}$ & \cellcolor[rgb]{0.761,0.851,0.766} 43.97$_{\pm0.16}$ & \cellcolor[rgb]{0.992,0.971,0.971} 40.19$_{\pm0.34}$ \\
\bottomrule
\end{tabular}
}
\caption{\textbf{LightMTP retains or even improves general LM benchmark performance --} perplexity on FineWeb-Edu and Lambada, and accuracy on Lambada and CORE, each result averaged over three seeds. FineWeb-Edu results report byte-level perplexity $2^{\text{bpb}}$ on a held-out shard. The Lambada columns show the FineWeb-Edu-pretrained \emph{base} model applied to Lambada's validation split, measuring perplexity and accuracy at the target word. In each column, the best value is \textbf{bolded} and the runner-up \underline{underlined}.}
\label{tab:d24_headline}
\end{table}

Prior MTP work found that language modeling performance degrades under MTP objectives. Our results confirm this for the three token-level MTP baselines: both Meta variants and DeepSeek MTP regress relative to NTP across every metric, with DeepSeek MTP showing the most severe degradation. Meta-L degrades less severely but still consistently underperforms NTP on every column.\\
\textbf{LightMTP, by contrast, shows almost no degradation and in several cases improves over NTP}. On FineWeb-Edu perplexity, LightMTP-C L16 matches NTP exactly, and most other LightMTP variants stay within $0.01$ of the NTP baseline. On CORE-avg, LightMTP-C with $k=10$ achieves the best score overall, with the default $k=4$ slightly behind NTP. LightMTP-S L16 ($k{=}10$) achieves the second-best Lambada perplexity in the table, only edged out by Future Summaries. This result aligns with Future Summaries' training on long-horizon targets obtained from their auxiliary right-to-left-trained target model. Notably, Future Summaries also avoids the degradation seen in the MTP baselines, though it trails both LightMTP variants as well as NTP on CORE-avg.\\
Although the variance-regularized (+VarReg) variants improved future-token anticipation in probing experiments, VarReg consistently harms general language modeling capabilities relative to their non-regularized LightMTP counterparts. Both +VarReg variants show worse FineWeb-Edu perplexity than their base counterpart, and the degradation continues on Lambada perplexity and accuracy. This divergence between probing-level gains and downstream degradation suggests gradient interference or representational competition between the future-token-anticipation objective and NTP, an explanation consistent with variance regularization sharpening the former at the latter's expense.

\begin{figure}[t]
\centering
\begin{minipage}[b]{0.48\textwidth}
  \vspace{0pt}
  \centering
  \small
\providecommand{\pkey}[1]{\textcolor[HTML]{#1}{\rule[0.15ex]{0.75em}{0.75em}}\hspace{0.35em}}%
\providecolor{gain}{HTML}{2E7D32}%
\begin{tabular}{lc}
\toprule
Objective & Held-out targets \\
\midrule
\pkey{8c8c8c}NTP & $46.4 \pm 1.4$ \\
\midrule
\phantom{\pkey{ffffff}}Meta, $k{=}4$ & \cellcolor{gain!29}$51.4 \pm 3.1$ \\
\pkey{a35f34}Meta, $k{=}8$ & \cellcolor{gain!66}$58.1 \pm 0.5$ \\
\midrule
\phantom{\pkey{ffffff}}Meta-L, $k{=}4$ & \cellcolor{gain!53}$55.8 \pm 0.9$ \\
\pkey{da8c62}Meta-L, $k{=}8$ & \cellcolor{gain!70}$58.9 \pm 1.2$ \\
\midrule
\phantom{\pkey{ffffff}}DeepSeek, $k{=}1$ & \cellcolor{gain!10}$47.8 \pm 2.5$ \\
\pkey{6da470}DeepSeek, $k{=}4$ & \cellcolor{gain!51}$55.4 \pm 0.4$ \\
\phantom{\pkey{ffffff}}DeepSeek, $k{=}8$ & \cellcolor{gain!50}$55.1 \pm 0.3$ \\
\midrule
\pkey{6e95c4}Future Summaries\textsuperscript{$\ddagger$} & $34.9 \pm 15.5$ \\
\midrule
\phantom{\pkey{ffffff}}LightMTP-C, $k{=}1$ & \cellcolor{gain!23}$50.1 \pm 2.6$ \\
\phantom{\pkey{ffffff}}LightMTP-C, $k{=}4$ & \cellcolor{gain!44}$54.0 \pm 3.0$ \\
\pkey{9985bc}LightMTP-C, $k{=}8$ & \cellcolor{gain!55}$56.1 \pm 1.4$ \\
\midrule
\phantom{\pkey{ffffff}}LightMTP-S, $k{=}1$ & \cellcolor{gain!25}$50.6 \pm 0.9$ \\
\phantom{\pkey{ffffff}}LightMTP-S, $k{=}4$ & \cellcolor{gain!52}$55.5 \pm 1.2$ \\
\pkey{d189a6}LightMTP-S, $k{=}8$ & \cellcolor{gain!58}$56.7 \pm 0.9$ \\
\bottomrule
\end{tabular}
\\$\ddagger$ held-out strict accuracy destabilizes late in training on 2 of 3 seeds (39.2 / 17.7 / 47.7).

  \captionof{table}{\textbf{MTP models better plan ahead -- } Countdown strict accuracy results on held-out targets (\%),
    mean $\pm$ s.d.\ over three seeds, greedy decoding. Shading reflects the gain over NTP.}
  \label{tab:countdown}
\end{minipage}\hfill
\begin{minipage}[b]{0.48\textwidth}
  \vspace{0pt}
  \centering
  \includegraphics{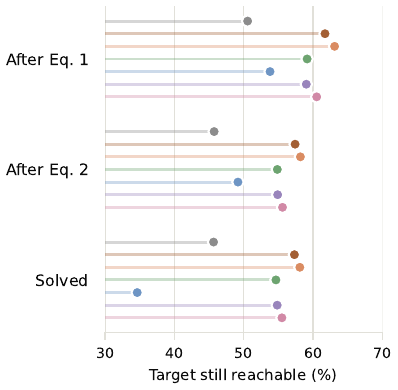}
  \captionof{figure}{\textbf{MTP retains more paths to the solution --} Plot of where the solution space is lost. Dots mark the share of held-out problems on which the target is still reachable after each equation. Colors match the table. Best $k$ per method.}
  \label{fig:countdown-analysis}
\end{minipage}
\end{figure}

\subsection{Planning: Countdown} \label{sec:planning}
We expect MTP training to improve planning, as it encourages models to consider longer horizons. In \emph{Countdown} \citep{gandhi2024stream}, players combine four numbers via basic arithmetic operators (+, -, ×, ÷) to arrive at a target number. Countdown requires models to explore the search space spanned by the numbers and operators, with which NTP has been shown to struggle.
We follow \citet{teoh_next-latent_2026} to generate data where targets $
\in [10, 100]$. We define accuracy as the ratio of arithmetically correct answers that consume each input number exactly once and end on the target.
\footnote{Note that since our models are built on top of the \texttt{nanochat} repository, the underlying transformer backbone is not identical to \citet{teoh_next-latent_2026} and so their numbers are not directly comparable to ours.}

On the held-out Countdown targets, \textbf{both LightMTP variants substantially outperform the NTP baseline and improve consistently as the lookahead horizon $k$ increases}. LightMTP-S outperforms LightMTP-C at every $k$ and does so with lower variance at the largest horizon. Notably, it achieves this with only $1.2$M MTP parameters, compared to $75$M for the Meta-L objective: a $98\%$ reduction in parameter count. Since for $k{=}1$ the Meta models are identical to an NTP model, they cannot offer gains at this horizon. Only Meta-L surpasses LightMTP at $k \geq 4$, reaching its own peak at $k{=}8$. The DeepSeek MTP objective peaks earlier, at $k{=}4$, then \emph{declines} at $k{=}8$, the only method for which additional lookahead is counterproductive. Despite its smaller footprint, LightMTP-S trails Meta's best result by only $2.2$ points ($56.7\%$ vs.\ $58.9\%$) while matching it in stability and monotonicity, suggesting a more favorable accuracy-per-parameter trade-off. The Future Summaries baseline, by contrast, underperforms NTP on average. However, this is due to held-out accuracy destabilizing late in training on two of three seeds. Together these results indicate that LightMTP extracts benefit from longer horizons more reliably than the DeepSeek and Future Summaries objectives, without the instability or non-monotonicity seen in these baselines, and with a fraction of Meta's parameters.

To test whether MTP models' gains reflect better planning, we examine the solution space remaining after each generated equation. Models that plan ahead should preserve more paths to the target, while greedy models are expected to more often enter dead ends. Figure~\ref{fig:countdown-analysis} shows the fraction of solutions remaining after each equation step. LightMTP-C and -S, as well as Meta and DeepSeek, preserve more solutions than NTP throughout the generation. This suggests that these \textbf{MTP methods do indeed plan ahead}: their first equations are not only arithmetically correct, but also tend to preserve a larger set of paths to the target. Future Summaries is an outlier: it preserves more solutions than NTP initially, but degrades over subsequent steps, resulting in lower final performance.

\subsection{Structure: GSM8K and MBPP}\label{sec:structure}

\begin{figure}[t]
\centering
\begin{minipage}[b]{0.46\textwidth}
  \vspace{0pt}
  \centering
  \resizebox{\linewidth}{!}{%
\begin{tabular}{lp{3.1em}p{3.1em}}
\toprule
\textbf{Method} & \begin{tabular}[b]{@{}c@{}}p@5\\relaxed\end{tabular} & \begin{tabular}[b]{@{}c@{}}p@5\\strict\end{tabular} \\
\midrule
NTP & \shortstack{41.97\\{\fontsize{4}{4}\selectfont $\pm$2.29}} & \shortstack{4.94\\{\fontsize{4}{4}\selectfont $\pm$1.44}} \\
Meta & \cellcolor[rgb]{0.973,0.897,0.897} \shortstack{41.45\\{\fontsize{4}{4}\selectfont $\pm$1.43}} & \cellcolor[rgb]{0.887,0.574,0.574} \shortstack{4.09\\{\fontsize{4}{4}\selectfont $\pm$0.08}} \\
Meta-L & \cellcolor[rgb]{0.997,0.988,0.988} \shortstack{41.91\\{\fontsize{4}{4}\selectfont $\pm$0.88}} & \cellcolor[rgb]{0.984,0.939,0.939} \shortstack{4.82\\{\fontsize{4}{4}\selectfont $\pm$1.19}} \\
DeepSeek MTP & \cellcolor[rgb]{0.855,0.452,0.452} \shortstack{39.17\\{\fontsize{4}{4}\selectfont $\pm$2.11}} & \cellcolor[rgb]{0.855,0.452,0.452} \shortstack{3.85\\{\fontsize{4}{4}\selectfont $\pm$0.68}} \\
Future Summaries & \cellcolor[rgb]{0.960,0.848,0.848} \shortstack{41.19\\{\fontsize{4}{4}\selectfont $\pm$0.53}} & \cellcolor[rgb]{0.995,0.997,0.995} \shortstack{4.95\\{\fontsize{4}{4}\selectfont $\pm$1.25}} \\
\midrule
LightMTP-C & \cellcolor[rgb]{0.687,0.805,0.693} \shortstack{\underline{43.62}\\{\fontsize{4}{4}\selectfont $\pm$1.51}} & \cellcolor[rgb]{0.983,0.989,0.983} \shortstack{\underline{4.98}\\{\fontsize{4}{4}\selectfont $\pm$0.91}} \\
LightMTP-S & \cellcolor[rgb]{0.649,0.781,0.655} \shortstack{\textbf{43.82}\\{\fontsize{4}{4}\selectfont $\pm$1.33}} & \cellcolor[rgb]{0.637,0.774,0.644} \shortstack{\textbf{5.69}\\{\fontsize{4}{4}\selectfont $\pm$1.09}} \\
\bottomrule
\end{tabular}
}

  \captionof{table}{\textbf{LightMTP is best on GSM8K -- } GSM8K results for NTP, MTP baselines and LightMTP. 1) ``relaxed'' considers whether the gold answer appears anywhere in the output. 2) ``strict'' only counts answers given in the correct answer format. Best result in \textbf{bold}, second-best \underline{underlined}.}
  \label{tab:gsm8k}
\end{minipage}\hfill
\begin{minipage}[b]{0.5\textwidth}
  \vspace{0pt}
  \centering
  \includegraphics[width=\linewidth]{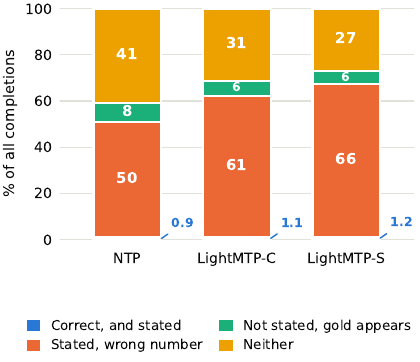}
  \captionof{figure}{\textbf{LightMTP better adheres to answer structure -- } The LightMTP models state the answer format more often than the NTP model, suggesting that LightMTP better picks up the format from the eight in-context examples.}
  \label{fig:gsm8k-analysis}
\end{minipage}
\end{figure}

Because MTP explicitly models immediate future tokens, we expect MTP models to exhibit better local look-ahead and better adherence to surface structure in generative tasks. We evaluate performance on GSM8K, a grade-school reasoning benchmark (\S~\ref{sec:GSM8K}), and MBPP, a Python coding dataset (\S~\ref{sec:MBPP}). We further analyze answer-format adherence and failure cases.

\paragraph{Reasoning on GSM8K.}
\label{sec:GSM8K}
GSM8K (Grade School Math 8K; \citealp{cobbe2021gsm8k}) is a benchmark evaluating models on math and reasoning. The examples require basic arithmetic and compositional skills.
We evaluate the base models only pretrained on FineWebEdu without additional finetuning on GSM8K's train set. At inference, we include eight train-set in-context examples to encourage the models to pick up the answer format. We present results in Table~\ref{tab:gsm8k}. The \emph{relaxed} metric checks whether the gold answer exists anywhere in the generated output, while the \emph{strict} metric only counts answers given in the format of "The answer is [ANSWER]" or ``\#\#\#\# [ANSWER]''. Since a model could potentially answer perfectly but without following the answer format we visualize the ratio of correct answers to format following in Figure~\ref{fig:gsm8k-analysis}.

On GSM8K, \textbf{LightMTP variants outperform NTP} on the relaxed metric across both configurations, \textbf{while the Meta and DeepSeek MTP baselines regress below NTP} on both metrics. 
Under the strict metric, where answers must follow the exact required format, LightMTP-S achieves the highest score, followed by LightMTP-C, essentially tied with NTP. Future Summaries again matches NTP-level performance. Both Meta variants and DeepSeek MTP underperform NTP on strict accuracy, despite being trained on 11.6\%, 20\% and 13\% more tokens, respectively. This reinforces that their regressions are not an artifact of the relaxed metric.

\begin{figure}[t]
\centering
\begin{minipage}[b]{0.45\textwidth}
  \vspace{0pt}
  \centering
  \small
  \resizebox{\linewidth}{!}{%
\begin{tabular}{lcccc}
\toprule
& \multicolumn{2}{c}{zero-shot} & \multicolumn{2}{c}{finetuned} \\
\cmidrule(lr){2-3} \cmidrule(lr){4-5}
\textbf{Method} & p@5 & p@10 & p@5 & p@10 \\
\midrule

NTP                                      & \shortstack{5.02\\{\fontsize{4}{4}\selectfont $\pm$0.97}} & \shortstack{8.52\\{\fontsize{4}{4}\selectfont $\pm$1.70}} & \shortstack{11.39\\{\fontsize{4}{4}\selectfont $\pm$1.22}} & \shortstack{15.19\\{\fontsize{4}{4}\selectfont $\pm$2.31}} \\
Meta & \cellcolor[rgb]{0.993,0.975,0.975} \shortstack{4.80\\{\fontsize{4}{4}\selectfont $\pm$1.80}} & \cellcolor[rgb]{0.974,0.900,0.900} \shortstack{7.04\\{\fontsize{4}{4}\selectfont $\pm$2.31}} & \cellcolor[rgb]{0.891,0.932,0.893} \shortstack{13.22\\{\fontsize{4}{4}\selectfont $\pm$2.64}} & \cellcolor[rgb]{0.880,0.925,0.882} \shortstack{17.78\\{\fontsize{4}{4}\selectfont $\pm$3.85}} \\
Meta-L  & \cellcolor[rgb]{0.725,0.829,0.731} \shortstack{6.01\\{\fontsize{4}{4}\selectfont $\pm$2.10}} & \cellcolor[rgb]{0.645,0.779,0.652} \shortstack{\underline{9.26}\\{\fontsize{4}{4}\selectfont $\pm$2.80}} & \cellcolor[rgb]{0.467,0.669,0.477} \shortstack{\textbf{16.43}\\{\fontsize{4}{4}\selectfont $\pm$0.48}} & \cellcolor[rgb]{0.564,0.729,0.572} \shortstack{\underline{21.85}\\{\fontsize{4}{4}\selectfont $\pm$0.64}} \\
DeepSeek                & \cellcolor[rgb]{0.855,0.452,0.452} \shortstack{0.19\\{\fontsize{4}{4}\selectfont $\pm$0.32}} & \cellcolor[rgb]{0.855,0.452,0.452} \shortstack{0.37\\{\fontsize{4}{4}\selectfont $\pm$0.64}} & \cellcolor[rgb]{0.855,0.452,0.452} \shortstack{2.47\\{\fontsize{4}{4}\selectfont $\pm$1.90}} & \cellcolor[rgb]{0.855,0.452,0.452} \shortstack{3.70\\{\fontsize{4}{4}\selectfont $\pm$2.57}} \\
Fut. Sum.         & \cellcolor[rgb]{0.571,0.733,0.580} \shortstack{\underline{6.57}\\{\fontsize{4}{4}\selectfont $\pm$0.63}} & \cellcolor[rgb]{0.822,0.890,0.826} \shortstack{8.89\\{\fontsize{4}{4}\selectfont $\pm$1.11}} & \cellcolor[rgb]{0.553,0.722,0.562} \shortstack{15.62\\{\fontsize{4}{4}\selectfont $\pm$1.14}} & \cellcolor[rgb]{0.613,0.759,0.620} \shortstack{21.11\\{\fontsize{4}{4}\selectfont $\pm$1.11}} \\
\midrule
LightMTP-C & \cellcolor[rgb]{0.467,0.669,0.477} \shortstack{\textbf{6.94}\\{\fontsize{4}{4}\selectfont $\pm$0.10}} & \cellcolor[rgb]{0.467,0.669,0.477} \shortstack{\textbf{9.63}\\{\fontsize{4}{4}\selectfont $\pm$0.64}} & \cellcolor[rgb]{0.468,0.669,0.478} \shortstack{\underline{16.42}\\{\fontsize{4}{4}\selectfont $\pm$1.18}} & \cellcolor[rgb]{0.467,0.669,0.477} \shortstack{\textbf{23.33}\\{\fontsize{4}{4}\selectfont $\pm$2.94}} \\
LightMTP-S             & \cellcolor[rgb]{0.924,0.952,0.925} \shortstack{5.30\\{\fontsize{4}{4}\selectfont $\pm$2.28}} & \cellcolor[rgb]{0.822,0.890,0.826} \shortstack{8.89\\{\fontsize{4}{4}\selectfont $\pm$4.44}} & \cellcolor[rgb]{0.877,0.924,0.879} \shortstack{12.55\\{\fontsize{4}{4}\selectfont $\pm$1.63}} & \cellcolor[rgb]{0.830,0.895,0.834} \shortstack{17.78\\{\fontsize{4}{4}\selectfont $\pm$2.22}} \\
\bottomrule
\end{tabular}
}
  \captionof{table}{\textbf{LightMTP-C outperforms on code generation --} Code-generation pass@$k$ (\%) for the d24 python-mix models (10\% Python in the pretraining mix) on MBPP. All models trained with $k=4$. Bold marks the best mean per column. All results reported as mean $\pm$ std over three seeds.}
  \label{tab:mbpp}
\end{minipage}\hfill
\begin{minipage}[b]{0.53\textwidth}
  \vspace{0pt}
  \centering
  \includegraphics[width=\linewidth]{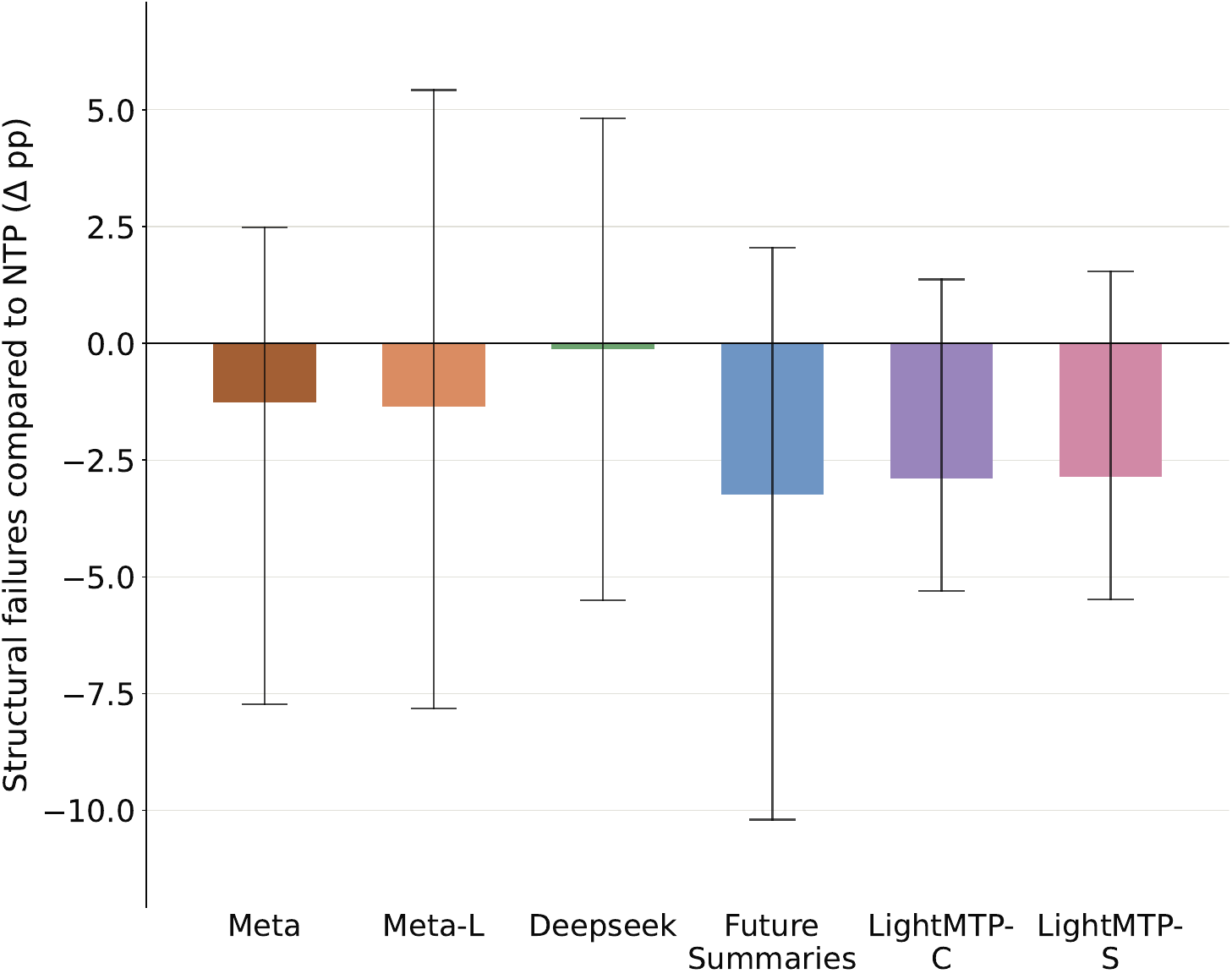}
  \captionof{figure}{\textbf{MTP reduces structural errors -- } Structural error rate of failed samples compared to NTP. Results over three seeds, with error bars indicating maximum and minimum scores. All MTP methods exhibit a lower structural error rate than NTP.}
  \label{fig:mbpp-structure}
\end{minipage}
\end{figure}

\paragraph{Coding on MBPP.}
\label{sec:MBPP}
Since code is more structured than natural language, we posit that it will especially benefit from MTP's improved local look-ahead capabilities.
Since FineWeb-Edu contains almost no code, we sample code data from The Stack \citep{kocetkov2022TheStack} V1.2's Python subset, and, for these experiments only, mix it into the pretraining dataloader so that about 10\% of the final data mixture is Python code. Then, we fine-tuned the pretrained models on MBPP's \citep{austin_program_2021} training set. We report results on the open validation set in Table~\ref{tab:mbpp}.

\textbf{In the zero-shot MBPP setting, LightMTP-C achieves the best performance at both pass@5 and pass@10, improving over NTP and all four MTP baselines}. After fine-tuning on MBPP, Meta-L takes the lead on pass@5, with LightMTP-C effectively tied and ahead on pass@10. LightMTP-S improves on NTP for finetuned MBPP pass@5 but performs well below LightMTP-C and Meta-L. DeepSeek MTP collapses on this task, scoring near zero in a zero-shot regime and remaining far behind even after finetuning, indicating that the objective transfers poorly to code.

To further investigate where MTP's edge comes from, we quantify the models' error types in Figure~\ref{fig:mbpp-structure}. Each failing completion was classified by the raised exception type. We report \emph{structural errors}, consisting of \texttt{NameError}, \texttt{SyntaxError}, \texttt{IndentationError}, and \texttt{ModuleNotFoundError}, i.e., failures that are attributable to local naming/syntax consistency rather than task or data understanding.

\textbf{Compared to NTP, all MTP models exhibit fewer structural errors}, to a minimal degree even DeepSeek, which underperformed NTP on task performance by a large margin. Fewer structural errors support our hypothesis that MTP objectives teach \emph{local} look-ahead skills, thus benefiting local coherence but not necessarily underlying task understanding. Future Summaries benefits from the largest reduction in structure-related errors, followed by the two LightMTP variants, and a more moderate effect on Meta and Meta-L. For some qualitative examples of structure-related generation differences between NTP and LightMTP, see Appendix~\ref{app:coding-examples}.

\section{Conclusion}

This paper introduces LightMTP, a latent multi-token prediction objective using a model's own future hidden states for supervision via a small auxiliary head. By predicting in latent space rather than over the vocabulary, it avoids the compute and memory overhead of token-level MTP, adding at most 1\% extra parameters and leaving the trunk and the inference path untouched. \textbf{Across probing, planning, reasoning, and code generation, LightMTP balances between anticipating future tokens and retaining next-token performance better} than all considered MTP baselines.

This balance is particularly relevant at smaller model scales ($<3B$), a regime for which prior work either reports degraded language modeling or that it omits entirely: LightMTP improves at modeling local task structure while leaving general language modeling benchmarks largely unaffected, making it practical as a pretraining objective where existing MTP methods are not. Future work characterizing the choices affecting the anticipation cost systematically, across target granularity, prediction horizon, and model scale, would provide a principled basis for designing multi-token objectives rather than selecting them empirically.

\section*{Ethics Statement}
This work proposes a pretraining objective for language models and evaluates it on publicly available data. It involves no human subjects, crowdsourcing, or personally collected data. We pretrain on FineWeb-Edu, a filtered subset of CommonCrawl, and, for the coding experiments, on the Python subset of The Stack v1.2, which contains only permissively licensed code and honors developer opt-out requests. We evaluate on established public benchmarks (CORE, LAMBADA, GSM8K, MBPP) and on synthetically generated Countdown problems. Models trained on web-scale corpora may inherit the biases, factual errors, and harmful content in that data. We did not apply safety tuning or bias mitigation beyond the filtering already performed by the dataset creators. Our models are small ($\leq$1.5B parameters), are not instruction-tuned, and were trained solely for research, so we consider their potential for misuse to be low. Training the models reported here required a substantial number of GPU hours, and we report all training budgets in Appendix~\ref{tab:param_count_complete} to support reproducibility.

\section*{AI Use Statement}
In this work, we used generative AI tools for code generation to support the implementation of our methods. We have not used generative AI tools for helping develop theoretical models or conceptual frameworks, designing or providing feedback on research methodology or experiments, supporting qualitative and thematic data analysis, or interpreting results, and the following are not applicable to this work: generating synthetic data, formulating mathematical claims, providing critical ingredients for proving mathematical claims sets, assisting in the writing of proofs, proposing or refining hypotheses, assisting with translation, cleaning and reformatting datasets. Additionally, we used generative AI tools for support on scientific figure creation, to help draft smaller paper sections and improve our paper's readability as well as identify relevant related work. We have reviewed all AI-assisted work via extensive testing (for coding), and a manual verification of paper and claim existence in the original papers (for paper writing and related work identification). We take responsibility for the final content of this work, including text, claims, or artifacts produced with the aid of generative AI.

\section*{Reproducibility}
All models in this paper are trained from scratch on publicly available data and built on top of \texttt{nanochat}, a publicly available codebase. The NTP base architecture, optimizer and hyperparameter settings all follow \texttt{nanochat}, without modifications. All MTP-specific alterations for every model variant and depth are fully described in the main text and in Table~\ref{tab:param_count_complete}. Our reimplementations of the MTP baselines, which deviate from the published architectures in the ways noted there, are described in Appendix~\ref{app:baseline-setup}. Appendix~\ref{app:exp-details} documents the preparation of every dataset we use and complete results. All data sources are public and all generated data is produced from fixed seeds. Every reported cell is a mean over three pretraining seeds, with standard deviations reported throughout.

\section*{Acknowledgments}
The project on which this publication is based was funded by the Federal Ministry of Research, Technology and Space under the funding code ``KI-Servicezentrum Berlin-Brandenburg'' 16IS22092. Responsibility for the content of this publication remains with the author.

\bibliography{iclr2027_conference}
\bibliographystyle{iclr2027_conference}
\newpage
\appendix
\section{Setup}
\label{app:setup}

\subsection{Architecture and Hyperparameters}
\label{sec:hyperparams}
All models build on the \texttt{nanochat} codebase, a compact reimplementation of a modern GPT-style decoder. We keep its architecture and training recipe unchanged across all methods and did not tune any hyperparameter for LightMTP or the baselines. Table~\ref{tab:backbone} lists the settings for the 24-layer models; the 12- and 18-layer models differ only in depth and in the depth-derived quantities (width, heads, weight decay, training steps).

\paragraph{Architecture.} Width scales with depth at a fixed aspect ratio ($d_\text{model} = 64 \cdot L$) with 128-dimensional attention heads. Each block is a pre-norm Transformer block with parameter-free RMSNorm, no bias terms, rotary position embeddings (base $10^4$) followed by QK-normalization, and a $4\times$ MLP with a squared-ReLU activation. The token embedding is normalized before the first block and the final hidden state before the output head. Input embedding and output head are untied, and output logits are soft-capped as $15\tanh(z/15)$. Attention alternates between local windows (``S'', half the context) and full context (``L'') in a repeating ``SSSL'' pattern, with the last layer always using full context. Two further additions are learnable per-layer scalars that rescale the residual stream and mix the normalized input embedding $x_0$ back in, $x \leftarrow \lambda^{\text{res}}_l x + \lambda^{0}_l x_0$, and ResFormer-style value embeddings: in every other layer (always including the last), a separate token-embedding table is added to the attention values through an input-dependent gate.

\paragraph{Optimization.} Following \texttt{nanochat}, all 2D weight matrices inside the Transformer blocks are optimized with Muon, using Nesterov momentum, Polar-Express orthogonalization (5 iterations), per-neuron second-moment normalization, and cautious weight decay. Embeddings, value embeddings, the output head and the per-layer scalars are optimized with AdamW. The learning rate is held constant for the first half of training and then decays linearly to zero. Muon momentum warms up from 0.85 to 0.95 over the first 300 steps, and weight decay decays linearly to zero over training. Training uses bfloat16 autocast (logits and loss in fp32), \texttt{torch.compile}, and FlashAttention-3 on Hopper GPUs. The auxiliary heads of all MTP methods are 2D matrices and are therefore optimized with Muon under the same settings as the trunk.

\paragraph{Data and tokenization.} We use \texttt{nanochat}'s byte-level BPE tokenizer with a vocabulary of $2^{15}{=}32{,}768$ tokens and a GPT-4-style pre-tokenization pattern (digits grouped in runs of at most two), trained on 2B characters of FineWeb-Edu. Sequences of 2,048 tokens are packed with a BOS-aligned best-fit strategy: every row starts at a document boundary and is filled without padding by the largest documents that fit, cropping one document to fill any remaining space. Attention is causal over the packed row, while the LightMTP targets are masked so that they never cross a document boundary (Section~\ref{sec:concat-target}).

\paragraph{Training budget.} Following \texttt{nanochat}'s speedrun recipe, the number of optimizer steps is set so that the number of training tokens is 12 times the number of scaling parameters (Transformer matrices plus output head; embeddings excluded). The auxiliary heads count as scaling parameters, so methods with larger heads train for more steps (Table~\ref{tab:param_count_complete}).

\begin{table}[h]
\centering
\small
\resizebox{\linewidth}{!}{%
\begin{tabular}{ll}
\toprule
\multicolumn{2}{l}{\textit{Architecture (d24)}} \\
Layers / width / heads & 24 / 1536 / 12 (head dim.\ 128, no GQA) \\
MLP & $4\times$ (6,144), ReLU$^2$, no biases \\
Normalization & parameter-free RMSNorm (pre-norm, embedding, final, QK) \\
Position encoding & RoPE, base $10^4$ \\
Attention windows & SSSL pattern (S = 1,024, L = 2,048 tokens), last layer L \\
Context length & 2,048 \\
Vocabulary & 32,768, untied input/output embeddings \\
Logit soft-cap & $15\tanh(z/15)$ \\
Extras & value embeddings (12 layers), per-layer $\lambda^\text{res}, \lambda^0$ scalars \\
\midrule
\multicolumn{2}{l}{\textit{Optimization}} \\
Batch size & $2^{19}{=}524{,}288$ tokens \\
Muon (block matrices) & Learning rate 0.02, momentum $0.85{\to}0.95$ (300 steps), Nesterov, \\
 & 5 Polar-Express steps, cautious weight decay $0.2\cdot(12/L)^2 = 0.05$ as linear decay \\
AdamW (embeddings, value emb.) & Learning rate $0.3\cdot(d_\text{model}/768)^{-1/2} \approx 0.212$ \\
AdamW (output head) & Learning rate $0.004\cdot(d_\text{model}/768)^{-1/2} \approx 0.0028$ \\
AdamW (scalars) & Learning rate 0.005 ($\lambda^\text{res}$), 0.5 ($\lambda^0$) \\
AdamW betas / eps / wd & (0.8, 0.95), $\lambda^0$: (0.96, 0.95) / $10^{-10}$ / 0 \\
LR schedule & no warmup; constant, then linear decay to 0 over the last 50\% \\
Gradient clipping & none \\
Precision & bfloat16 autocast, fp32 logits and loss \\
Token budget & $12\times$ scaling parameters (Table~\ref{tab:param_count_complete}) \\
\bottomrule
\end{tabular}
}
\caption{Backbone architecture and training hyperparameters for the 24-layer models, inherited unchanged from \texttt{nanochat}.}
\label{tab:backbone}
\end{table}

\begin{table}[h!]
\centering
\small
\setlength{\tabcolsep}{5pt}
\resizebox{\linewidth}{!}{%
\begin{tabular}{lcccccrrr}
\toprule
Model & Total & Scaling & Non-scaling & MTP head & $\Delta$ params & Tokens & Steps & H100 GPU-h \\
\midrule
\multicolumn{9}{@{}l}{\textbf{d12}\quad $768\times12\times6$}\\
\quad NTP & 286.3M & 110.1M & 176.2M & --- & --- & 1.32B & 2\,520 & 0.7 \\
\quad Meta-L$^{\ddagger}$ & 361.8M & 185.6M & 176.2M & 75.5M & +26.4\% & 2.23B & 4\,248 & 2.0 \\
\quad Meta & 307.5M & 131.3M & 176.2M & 21.2M & +7.4\% & 1.58B & 3\,006 & 1.6 \\
\quad DeepSeek MTP & 311.0M & 134.9M & 176.2M & 24.8M & +8.7\% & 1.62B & 3\,087 & 1.6 \\
\quad Future Summaries & 287.4M(+286.3M) & 111.3M & 176.2M & 1.2M & +0.4\% & 1.34B & 2\,547 & 0.9 \\
\quad LightMTP-single & 287.4M & 111.3M & 176.2M & 1.2M & +0.4\% & 1.34B & 2\,547 & 0.7 \\
\quad LightMTP-concat & 289.2M & 113.1M & 176.2M & 2.9M & +1.0\% & 1.36B & 2\,587 & 0.8 \\
\midrule
\multicolumn{9}{@{}l}{\textbf{d18}\quad $1152\times18\times9$}\\
\quad NTP & 701.9M & 324.4M & 377.5M & --- & --- & 3.89B & 7\,425 & 5.4 \\
\quad Meta-L$^{\ddagger}$ & 815.1M & 437.7M & 377.5M & 113.2M & +16.1\% & 5.25B & 10\,017 & 9.6 \\
\quad Meta & 749.7M & 372.2M & 377.5M & 47.8M & +6.8\% & 4.47B & 8\,518 & 9.1 \\
\quad DeepSeek MTP & 757.6M & 380.2M & 377.5M & 55.7M & +7.9\% & 4.56B & 8\,700 & 9.3 \\
\quad Future Summaries & 704.5M(+701.9M) & 327.1M & 377.5M & 2.7M & +0.4\% & 3.92B & 7\,485 & 7.0 \\
\quad LightMTP-single & 704.5M & 327.1M & 377.5M & 2.7M & +0.4\% & 3.92B & 7\,485 & 5.5 \\
\quad LightMTP-concat & 708.5M & 331.0M & 377.5M & 6.6M & +0.9\% & 3.97B & 7\,576 & 5.7 \\
\midrule
\multicolumn{9}{@{}l}{\textbf{d24}\quad $1536\times24\times12$}\\
\quad NTP & 1.38B & 729.8M & 654.3M & --- & --- & 8.76B & 16\,704 & 24.5 \\
\quad Meta-L$^{\ddagger}$ & 1.54B & 880.8M & 654.3M & 151.0M & +10.9\% & 10.57B & 20\,160 & 37.9 \\
\quad Meta & 1.47B & 814.7M & 654.3M & 84.9M & +6.1\% & 9.78B & 18\,648 & 37.9 \\
\quad DeepSeek MTP & 1.48B & 828.9M & 654.3M & 99.1M & +7.2\% & 9.95B & 18\,972 & 37.4 \\
\quad Future Summaries & 1.39B(+1.38B) & 734.5M & 654.3M & 4.7M & +0.3\% & 8.81B & 16\,812 & 32.3 \\
\quad LightMTP-single & 1.39B & 734.5M & 654.3M & 4.7M & +0.3\% & 8.81B & 16\,812 & 25.6 \\
\quad LightMTP-concat & 1.40B & 741.6M & 654.3M & 11.8M & +0.9\% & 8.90B & 16\,974 & 26.2 \\
\bottomrule
\end{tabular}
}
\caption{Scaling and non-scaling (token and value embeddings) parameter counts for models with 12, 18, and 24 transformer layers. Training tokens and optimizer steps for each
architecture. ``MTP head'' is the auxiliary head(s) each variant adds on top of the NTP model. ``$\Delta$ params'' denotes the percentage of extra parameters the method adds compared to just NTP. Token and step budgets follow nanochat's \texttt{speedrun.sh}: a data:parameter ratio of $12$ over $n_\text{matrix}+n_\text{head}$ (transformer matrices plus output head, excluding token and value embeddings). Because the ratio is applied to a parameter count that grows with the MTP head size, variants that add more parameters are trained for proportionally more steps. Future Summaries requires an already right-to-left-trained, NTP-based helper model to generate informative latent targets. ``H100 GPU-h'' is the approximate pretraining cost in GPU-hours on one NVIDIA H100.}
\label{tab:param_count_complete}
\end{table}

\subsection{Baselines Setup}
\label{app:baseline-setup}
To make the comparison between LightMTP and the baseline MTP approaches fairer, we adjusted their implementations from the original works.

We report two token-level MTP baselines derived from \citet{gloeckle_better_2024}, denoted \textbf{Meta} and \textbf{Meta-L}. Both follow the same recipe: a shared transformer trunk produces a hidden state at position $t$, from which $k$ heads predict the tokens $x_{t+1}, \dots, x_{t+k}$. They differ only in what a head \emph{is}.

\paragraph{Meta (default).}
Each of the $k{-}1$ auxiliary heads is a single transformer layer applied to the trunk's final hidden state, followed by the shared unembedding, exactly as in the architecture
\citet{gloeckle_better_2024} propose in the main body of their paper. We deviate from their design in one aspect: the $k{=}1$ head is a \emph{linear projection only}, i.e., the shared unembedding matrix is applied directly to the final hidden state, rather than having a transformer layer of its own. This keeps the next-token path identical to that of NTP and of every other model we compare against, so the next-token capacity is the same across the table and any difference we measure is attributable to the auxiliary objective rather than to extra depth on the next-token path.

\paragraph{Meta-L.}
The $k{-}1$ auxiliary heads are untied \emph{linear} LM heads, each an independent copy of the
unembedding matrix, reading the same final hidden state. This variant is also presented in
\citet{gloeckle_better_2024}, in the alternative-architecture ablation of their appendix, where heads that are a single linear layer without nonlinearity are compared against the
transformer-layer heads of the main architecture. The authors report no significant improvement from the alternatives. We keep it as a separate baseline because it isolates the effect of separate unembedding matrices.

\paragraph{DeepSeek-v3-style MTP.} DeepSeek-V3 \citep{deepseek-ai_deepseek-v3_2025} augments next-token prediction with $D=k-1$ sequential MTP modules that preserve a causal chain across the predicted positions. Each module applies RMSNorm to the previous depth's representation $\vh_i^{k-1}$ and to the embedding of token $t_{i+k}$, concatenates the two, projects the result with $M_k\in\mathbb{R}^{d\times 2d}$, and passes it through its own Transformer block. The embedding table and the output head are shared with the main model, so each module adds only $M_k$ and one block. The per-depth cross-entropy losses are averaged and added to the NTP loss with weight $\lambda$. We keep this structure and adapt two settings to our comparison. First, we use $D=k-1=3$ modules rather than the $D=1$ of DeepSeek-V3, so that the objective supervises the same four positions as the other methods. Second, we set $\lambda=1$, as for every other method, rather than DeepSeek-V3's schedule of $0.3$ decaying to $0.1$, since we do not tune loss weights per method. Each module's transformer block is randomly initialized and inherits the local attention window of our backbone's first layer (1,024 tokens) rather than attending over the full context. As in the original, the modules are discarded at inference. At depth 24 they add 99M parameters (Table~\ref{tab:param_count_complete}).

\paragraph{Future Summaries.} Future Summary Prediction \citep{mahajan_beyond_2026} trains an auxiliary head to predict a compact representation of the entire future continuation rather than individual future tokens. We reproduce their learned-summary variant: a separate reverse language model is pretrained right-to-left on the same corpus, and the left-to-right model is then trained to predict that model's hidden state from its own. The reverse model is frozen throughout and is never used at inference. As in the original, we treat its pretraining as a one-time overhead and mostly exclude it from the token budgets in Table~\ref{tab:param_count_complete}. It has the same depth, width and vocabulary as the model it supervises, so it roughly doubles the total pretraining compute of this baseline. We deviate in three aspects. First, we implement only the learned-summary variant, not the handcrafted bag-of-words summaries the original also proposes. Second, the auxiliary head is the same two-layer MLP with a GELU nonlinearity that LightMTP uses, rather than the transformer block of the original, so that the two latent methods are compared at equal head capacity. Third, we use a cosine loss on unit-normalized vectors rather than the original's $\ell_2$ loss, again matching LightMTP. Targets are taken from the reverse model's final layer and predicted from the final layer of the left-to-right trunk. At depth 24 the head adds 4.7M parameters, the same as LightMTP single-target.

\section{Further Experiment Details}
\label{app:exp-details}
\subsection{Future token anticipation}
\label{app:probing}
Linear probes are trained on hidden states extracted from FineWeb-Edu shard 435, which is neither in the training nor in the validation split of any model we report. Probe positions whose target token is a special token are discarded, matching the masking used for the pretraining bits-per-byte metric. For each horizon $k$, positions are subsampled to at most 2M and then split 80/20 into probe-train and probe-test sets, using a fixed permutation seed (42) shared across all models, so every model is probed on exactly the same positions.
\begin{figure}[h!]
    \centering
    \includegraphics[width=\linewidth]{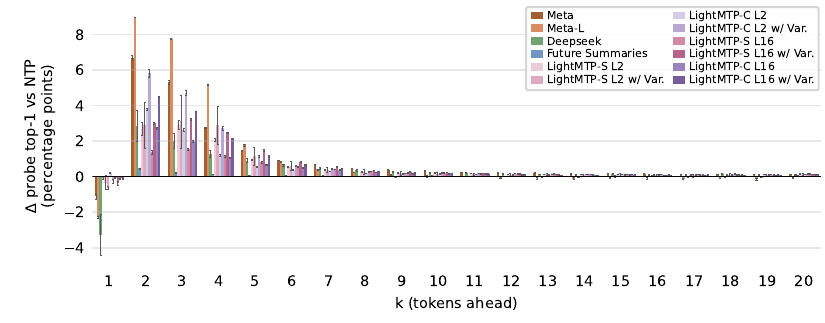}
    \caption{Probing top-1 accuracy on the models' frozen last hidden states over three seeds for $k=1, \dots,  20$. All MTP models exhibit higher future token anticipation than NTP, but this often comes with a performance trade-off at $k=1$. While for $2\leq k \leq 5$, Meta's MTP-L exhibits the strongest anticipation effect, its anticipation then gradually anneals to the other MTP models' performance until it dips \emph{below} NTP for $k \geq 12$. LightMTP models anticipation effect for early $k$ is moderate but it benefits from a mitigated degradation at $k=1$.}
    \label{fig:probing-complete}
\end{figure}

\subsection{LM Benchmarks}
CORE and LAMBADA are evaluated from the DCLM-derived evaluation bundle shipped with \texttt{nanochat}, using its per-task few-shot counts and prompt templates unchanged. Few-shot examples are drawn so all models see identical prompts.
MMLU \citep{hendrycks2021measuring} is evaluated as a 10-shot multiple-choice task over all 14{,}042 test questions from 57 categories. Each question comes with its four options already inserted into the prompt. We score by comparing the model's likelihood of each answer letter. Due to the models' limited pretraining budget and modest sizes, their MMLU results are not distinguishable from chance level (25\%) and therefore only reported for completeness's sake in \ref{tab:d24_preliminary_short} in the Appendix.

\subsection{Countdown}
Countdown problems are generated following \citet{teoh_next-latent_2026}: We reserve 10\% of the \emph{targets} for out-of-distribution evaluation, so held-out problems have targets the model never saw during training. We generate 500{,}000 training problems, 10{,}000 in-distribution, and 10{,}000 held-out-target evaluation problems, each with four input numbers and a three-equation solution. The in-distribution evaluation set is deduplicated against the training set by exact string match, the held-out-target set is distinct by construction. Every generated solution is verified to be arithmetically valid and to reach its target. Countdown uses its own tokenizer, in which each integer below 10{,}000 is a single token, and eight pause tokens \citep{goyal_think_2024} separate the problem from the solution.

\subsection{GSM8K}
For GSM8K, we evaluate on the 1{,}319-example test split, with eight in-context examples sampled per test item from the official 7{,}473-example training split. Exemplars are rendered as \texttt{Q: \{question\}\textbackslash nA: \{rationale\} The answer is \{answer\}.}, and examples are gradually dropped if a prompt exceeds the context window.

\subsection{MBPP}
MBPP uses the official task-ID splits: tasks 601--974 for finetuning, tasks 511--600 for evaluation, and tasks 2--4 from the reserved prompt split as few-shot exemplars, so the exemplars never overlap the evaluation set. Prompts follow the original MBPP format (problem statement, the three assert statements, and a \texttt{[BEGIN]}/\texttt{[DONE]} delimited solution). Generated code is cut at \texttt{[DONE]} and executed in a sandbox against the problem's asserts, with a five-second timeout. During finetuning, the loss is masked over the prompt and applied only to the solution.
\begin{table}[h]
\centering
\small
\setlength{\tabcolsep}{4pt}
\resizebox{\linewidth}{!}{%
\begin{tabular}{lcccccc}
\toprule
& NTP & Meta & Meta-L & \shortstack{Future\\Summaries} & \shortstack{LightMTP-C} & \shortstack{LightMTP-S} \\
\midrule
\multicolumn{7}{l}{\textit{Over all completions}} \\
\quad parses as Python            & 96.00{\fontsize{6}{6}\selectfont $\pm$1.11} & \textbf{96.44}{\fontsize{6}{6}\selectfont $\pm$1.16} & 96.07{\fontsize{6}{6}\selectfont $\pm$1.01} & 95.33{\fontsize{6}{6}\selectfont $\pm$1.49} & 96.41{\fontsize{6}{6}\selectfont $\pm$1.43} & 95.70{\fontsize{6}{6}\selectfont $\pm$0.17} \\
\quad solved                      & 4.07{\fontsize{6}{6}\selectfont $\pm$0.56} & 4.48{\fontsize{6}{6}\selectfont $\pm$1.12} & \textbf{6.37}{\fontsize{6}{6}\selectfont $\pm$0.17} & 5.44{\fontsize{6}{6}\selectfont $\pm$0.97} & 5.70{\fontsize{6}{6}\selectfont $\pm$0.53} & 4.00{\fontsize{6}{6}\selectfont $\pm$0.29} \\
\midrule
\multicolumn{7}{l}{\textit{Structural failures (\% of failing completions)}} \\
\quad NameError                   & 8.27{\fontsize{6}{6}\selectfont $\pm$3.66} & 7.93{\fontsize{6}{6}\selectfont $\pm$2.37} & 7.16{\fontsize{6}{6}\selectfont $\pm$0.44} & \textbf{5.21}{\fontsize{6}{6}\selectfont $\pm$0.37} & 6.36{\fontsize{6}{6}\selectfont $\pm$1.11} & 5.90{\fontsize{6}{6}\selectfont $\pm$1.41} \\
\quad SyntaxError                 & 4.13{\fontsize{6}{6}\selectfont $\pm$0.99} & 3.68{\fontsize{6}{6}\selectfont $\pm$0.59} & 3.72{\fontsize{6}{6}\selectfont $\pm$0.65} & 4.61{\fontsize{6}{6}\selectfont $\pm$1.34} & \textbf{3.66}{\fontsize{6}{6}\selectfont $\pm$1.29} & 4.20{\fontsize{6}{6}\selectfont $\pm$1.53} \\
\quad IndentationError            & 0.93{\fontsize{6}{6}\selectfont $\pm$0.91} & \textbf{0.51}{\fontsize{6}{6}\selectfont $\pm$0.30} & 0.71{\fontsize{6}{6}\selectfont $\pm$0.36} & 0.63{\fontsize{6}{6}\selectfont $\pm$0.64} & 0.55{\fontsize{6}{6}\selectfont $\pm$0.67} & 0.81{\fontsize{6}{6}\selectfont $\pm$0.81} \\
\quad ModuleNotFoundError         & 0.43{\fontsize{6}{6}\selectfont $\pm$0.24} & 0.39{\fontsize{6}{6}\selectfont $\pm$0.57} & 0.83{\fontsize{6}{6}\selectfont $\pm$0.31} & 0.08{\fontsize{6}{6}\selectfont $\pm$0.07} & 0.31{\fontsize{6}{6}\selectfont $\pm$0.55} & \textbf{0.00}{\fontsize{6}{6}\selectfont $\pm$0.00} \\
\quad \textit{Structural (total)}   & \textit{13.76}{\fontsize{6}{6}\selectfont $\pm$5.51} & \textit{12.51}{\fontsize{6}{6}\selectfont $\pm$2.66} & \textit{12.42}{\fontsize{6}{6}\selectfont $\pm$1.11} & \textit{\textbf{10.53}}{\fontsize{6}{6}\selectfont $\pm$1.83} & \textit{10.88}{\fontsize{6}{6}\selectfont $\pm$2.54} & \textit{10.92}{\fontsize{6}{6}\selectfont $\pm$2.35} \\
\midrule
\multicolumn{7}{l}{\textit{Non-structural, for contrast (\% of failing)}} \\
\quad AssertionError (ran, wrong answer)& \textbf{58.84}{\fontsize{6}{6}\selectfont $\pm$1.83} & 62.01{\fontsize{6}{6}\selectfont $\pm$3.06} & 63.21{\fontsize{6}{6}\selectfont $\pm$0.96} & 61.66{\fontsize{6}{6}\selectfont $\pm$1.38} & 59.74{\fontsize{6}{6}\selectfont $\pm$5.43} & 59.91{\fontsize{6}{6}\selectfont $\pm$2.22} \\
\quad TypeError                   & 17.83{\fontsize{6}{6}\selectfont $\pm$1.70} & 16.49{\fontsize{6}{6}\selectfont $\pm$3.96} & \textbf{13.93}{\fontsize{6}{6}\selectfont $\pm$1.34} & 17.42{\fontsize{6}{6}\selectfont $\pm$2.07} & 20.58{\fontsize{6}{6}\selectfont $\pm$1.93} & 20.18{\fontsize{6}{6}\selectfont $\pm$0.96} \\
\quad IndexError                  & 4.32{\fontsize{6}{6}\selectfont $\pm$1.63} & 3.95{\fontsize{6}{6}\selectfont $\pm$0.58} & 4.31{\fontsize{6}{6}\selectfont $\pm$0.39} & 5.06{\fontsize{6}{6}\selectfont $\pm$1.38} & \textbf{3.18}{\fontsize{6}{6}\selectfont $\pm$0.83} & 3.82{\fontsize{6}{6}\selectfont $\pm$0.72} \\
\quad AttributeError              & 2.55{\fontsize{6}{6}\selectfont $\pm$0.52} & 2.21{\fontsize{6}{6}\selectfont $\pm$0.80} & 2.33{\fontsize{6}{6}\selectfont $\pm$0.91} & 2.86{\fontsize{6}{6}\selectfont $\pm$0.36} & 2.83{\fontsize{6}{6}\selectfont $\pm$1.12} & \textbf{2.16}{\fontsize{6}{6}\selectfont $\pm$1.36} \\
\quad Timeout                     & 0.73{\fontsize{6}{6}\selectfont $\pm$0.34} & 1.12{\fontsize{6}{6}\selectfont $\pm$0.23} & 1.07{\fontsize{6}{6}\selectfont $\pm$0.85} & \textbf{0.51}{\fontsize{6}{6}\selectfont $\pm$0.07} & 0.75{\fontsize{6}{6}\selectfont $\pm$0.36} & 0.73{\fontsize{6}{6}\selectfont $\pm$0.52} \\
\bottomrule
\end{tabular}
}
\caption{MBPP failure-mode breakdown, on the MBPP validation split (tasks 511--600): 90 problems $\times$ 10 samples $=$ 900 completions per model/seed, temperature $0.7$, mean $\pm$ std over seeds 42, 7, and 1337. \emph{Parses} and \emph{solved} are percentages of \emph{all} completions. Each failure row reports a percentage of that model's \emph{failing} completions, re-executed against the problem's asserts and labeled by the exception raised. \emph{Structural} sums the four exception types attributable to local naming and syntax consistency rather than task or data understanding.}
\label{tab:mbpp_structural_full}
\end{table}

\subsection{MBPP: Qualitative Examples}
\label{app:coding-examples}
We provide two qualitative examples illustrating common structural failures NTP is more prone to than LightMTP.
    
\begin{lstlisting}[language=Python]
def find_Min_Sum(n): 
    result = 0
    for i in range(2,int(math.sqrt(n)) + 1):  # uses math.sqrt, never imported "math"
        ...

def modular_sum(nums, m, divisor):
    if (dividor % m == 0):   # defined "divisor", referenced "dividor" a few tokens later
\end{lstlisting}

\newpage
\subsection{What MTP horizon works best?}
\begin{figure}[h]
  \centering
  \captionsetup[subfloat]{justification=justified, singlelinecheck=false}
  \begin{minipage}[t]{0.49\textwidth}
    \vspace{0pt}
    \subfloat[Single-target: linear-probe top-1 gain over NTP for decoding $x_{t+j}$ from the final-layer state at position $t$. Results over three seeds. Shade encodes $k$ Bands: 95\% CI.\label{fig:ksweep:single}]{%
      \includegraphics[width=\linewidth]{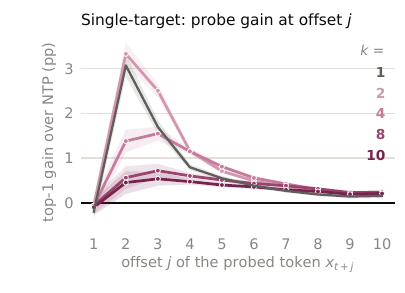}}
  \end{minipage}\hfill
  \begin{minipage}[t]{0.49\textwidth}
    \vspace{0pt}
    \subfloat[Concat: as in \protect\ref{fig:ksweep:single}, for the concatenated next-$k$ targets.\label{fig:ksweep:concat}]{%
      \includegraphics[width=\linewidth]{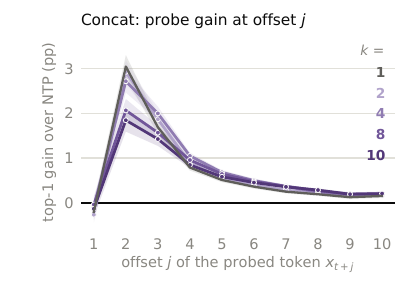}}
  \end{minipage}
\medskip
  \includegraphics[width=\textwidth]{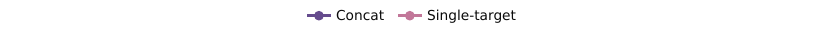}
  \medskip
  \begin{minipage}[t]{0.49\textwidth}
    \vspace{0pt}
    \subfloat[Anticipation cost per $k$ computed over $t{+}2..t{+}10$; lower is better. Dots: per-seed costs; bars: 95\% CI.\label{fig:ksweep:cost}]{%
      \includegraphics[width=\linewidth]{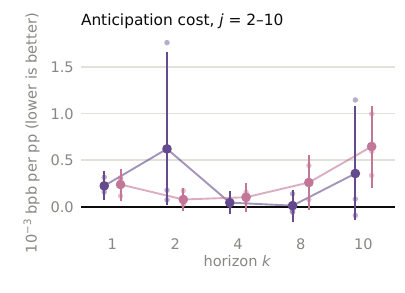}}
  \end{minipage}\hfill
  \begin{minipage}[t]{0.49\textwidth}
    \vspace{0pt}
    \subfloat[CORE average over 22 tasks, difference from NTP. Dots: per-seed differences. Bars: 95\% CI.\label{fig:ksweep:core}]{%
      \includegraphics[width=\linewidth]{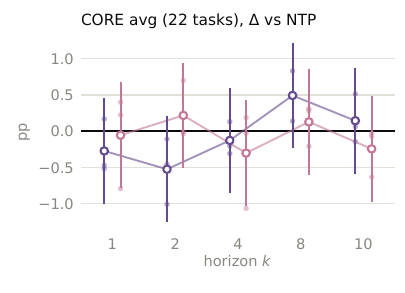}}
  \end{minipage}

  \caption{Effect of the prediction horizon $k$ for LightMTP at target layer 16 (head layer 23), relative to NTP. NTP uses 5 seeds and every other configuration 3. All sweep runs use target layer 16 without VarReg ($k{=}4$ is the LightMTP L16 configuration).}
  \label{fig:ksweep}
\end{figure}

For both LightMTP variants, we run a sweep over the MTP horizon $k \in \{1, 2, 4, 8, 10\}$. Future token anticipation changes and differences in CORE results are visualized in the panels of Figure~\ref{fig:ksweep}.
\label{sec:horizon-k}
\begin{landscape}
\subsubsection{Full Horizon Sweep Results}
    \begin{table}[!b]
\centering
\tiny
\setlength{\tabcolsep}{2pt}
\resizebox{18cm}{!}{%
\begin{tabular}{lp{2.2em}p{2.2em}p{2.2em}p{2.2em}p{2.2em}p{2.2em}p{2.2em}p{2.2em}p{2.2em}p{2.2em}p{2.2em}p{2.2em}p{2.2em}p{2.2em}p{2.2em}p{2.2em}p{2.2em}p{2.2em}p{2.2em}p{2.2em}p{2.2em}p{2.2em}p{2.2em}p{2.2em}p{2.2em}p{2.2em}p{2.2em}}
\toprule
 & \multicolumn{6}{c}{\textbf{World Knowledge}} & \multicolumn{7}{c}{\textbf{Commonsense}} & \multicolumn{4}{c}{\textbf{Passage Comprehension}} & \multicolumn{5}{c}{\textbf{Symbolic reasoning}} &  &  &  &  &  \\
\cmidrule(lr){2-7}\cmidrule(lr){8-14}\cmidrule(lr){15-18}\cmidrule(lr){19-23}
\textbf{Method} & \rotatebox{60}{\textbf{jeopardy}} & \rotatebox{60}{\textbf{bb-qa-wikidata}} & \rotatebox{60}{\textbf{arc-easy}} & \rotatebox{60}{\textbf{arc-challenge}} & \rotatebox{60}{\textbf{openbook-qa}} & \rotatebox{60}{\textbf{bb-language-identification}} & \rotatebox{60}{\textbf{copa}} & \rotatebox{60}{\textbf{commonsense-qa}} & \rotatebox{60}{\textbf{piqa}} & \rotatebox{60}{\textbf{hellaswag}} & \rotatebox{60}{\textbf{hellaswag-zeroshot}} & \rotatebox{60}{\textbf{winograd}} & \rotatebox{60}{\textbf{winogrande}} & \rotatebox{60}{\textbf{squad}} & \rotatebox{60}{\textbf{coqa}} & \rotatebox{60}{\textbf{boolq}} & \rotatebox{60}{\textbf{lambada-openai}} & \rotatebox{60}{\textbf{bb-dyck-languages}} & \rotatebox{60}{\textbf{bb-cs-algorithms}} & \rotatebox{60}{\textbf{bb-repeat-copy-logic}} & \rotatebox{60}{\textbf{bb-operators}} & \rotatebox{60}{\textbf{lsat-ar}} & \rotatebox{60}{\textbf{CORE avg}} & \rotatebox{60}{\textbf{MMLU (ICL)}} & \rotatebox{60}{\textbf{GSM8K (p@5)}} & \rotatebox{60}{\textbf{GSM8K (strict)}} & \rotatebox{60}{\textbf{Lambada (PPL)}} \\
\midrule
\multicolumn{28}{l}{\textit{Baseline}} \\
\bottomrule
\end{tabular}
}
\caption{depth-24 horizon sweep with target layer 16: latent MTP with concatenated next-$k$ hidden-state targets (Concat) or the single hidden state at $t{+}k$ (Single-target), for $k\in\{1,2,4,8,10\}$, against NTP. NTP is averaged over five seeds, as in the main results tables; every other row is three seeds. At $k{=}1$, the two objectives coincide. Full CORE suite: per-task in-context accuracy (\%), grouped into closed-book knowledge, everyday commonsense, passage comprehension, and symbolic reasoning, followed by the CORE average over its 22 tasks, and MMLU and GSM8K pass@5, which are reported outside the CORE average. Cells color-coded per column, green (improvement) to red (regression) vs NTP; in each column the best value is \textbf{bolded} and the runner-up \underline{underlined}. CORE (centered) averages each of the 22 CORE tasks as $(\mathrm{acc}-b)/(1-b)$ with respect to random baseline $b$.}
\label{tab:d24_ksweep_core}
\end{table}

\end{landscape}

\begin{landscape}
    \subsection{Full LM benchmark results}
    \begin{table}[h]
\centering
\small
\setlength{\tabcolsep}{2pt}
\resizebox{17cm}{!}{%
\begin{tabular}{lp{2.2em}p{2.2em}p{2.2em}p{2.2em}p{2.2em}p{2.2em}p{2.2em}p{2.2em}p{2.2em}p{2.2em}p{2.2em}p{2.2em}p{2.2em}p{2.2em}p{2.2em}p{2.2em}p{2.2em}p{2.2em}p{2.2em}p{2.2em}p{2.2em}p{2.2em}p{2.2em}p{2.2em}p{2.2em}p{2.2em}p{2.2em}}
\toprule
 & \multicolumn{6}{c}{\textbf{World Knowledge}} & \multicolumn{7}{c}{\textbf{Commonsense}} & \multicolumn{4}{c}{\textbf{Passage Comprehension}} & \multicolumn{5}{c}{\textbf{Symbolic reasoning}} &  &  &  &  &  \\
\cmidrule(lr){2-7}\cmidrule(lr){8-14}\cmidrule(lr){15-18}\cmidrule(lr){19-23}
\textbf{Method} & \rotatebox{60}{\textbf{jeopardy}} & \rotatebox{60}{\textbf{bb-qa-wikidata}} & \rotatebox{60}{\textbf{arc-easy}} & \rotatebox{60}{\textbf{arc-challenge}} & \rotatebox{60}{\textbf{openbook-qa}} & \rotatebox{60}{\textbf{bb-language-identification}} & \rotatebox{60}{\textbf{copa}} & \rotatebox{60}{\textbf{commonsense-qa}} & \rotatebox{60}{\textbf{piqa}} & \rotatebox{60}{\textbf{hellaswag}} & \rotatebox{60}{\textbf{hellaswag-zeroshot}} & \rotatebox{60}{\textbf{winograd}} & \rotatebox{60}{\textbf{winogrande}} & \rotatebox{60}{\textbf{squad}} & \rotatebox{60}{\textbf{coqa}} & \rotatebox{60}{\textbf{boolq}} & \rotatebox{60}{\textbf{lambada-openai}} & \rotatebox{60}{\textbf{bb-dyck-languages}} & \rotatebox{60}{\textbf{bb-cs-algorithms}} & \rotatebox{60}{\textbf{bb-repeat-copy-logic}} & \rotatebox{60}{\textbf{bb-operators}} & \rotatebox{60}{\textbf{lsat-ar}} & \rotatebox{60}{\textbf{CORE avg}} & \rotatebox{60}{\textbf{MMLU (ICL)}} & \rotatebox{60}{\textbf{GSM8K (p@5)}} & \rotatebox{60}{\textbf{GSM8K (strict)}} & \rotatebox{60}{\textbf{Lambada (PPL)}} \\
\midrule
\multicolumn{28}{l}{\textit{Baselines}} \\
\bottomrule
\end{tabular}
}
\ifdefined\captionsetup\captionsetup{font=footnotesize}\fi
\caption{Full depth-24 language modeling benchmark results: full CORE benchmark suite (per-task ICL accuracy), CORE average, MMLU (ICL), GSM8K pass@5 under strict and relaxed matchers and Lambada perplexity and accuracy; mean $\pm$ std over three seeds for every row, five for NTP. Every MTP model uses $k{=}4$ targets. Every run is trained with parameter-matched budget, so the number of steps follows each model's parameter count. Meta-L trains on 20.7\%, DeepSeek MTP on 13.6\%, and Meta on 11.6\% more tokens than NTP. CORE (centered) averages each of the 22 CORE tasks as $(\mathrm{acc}-b)/(1-b)$ with respect to random baseline $b$. \textbf{LAMBADA columns.} The base model (not a Lambada-finetuned variant) on the LAMBADA test split: target-word perplexity (lower is better) and accuracy.}
\label{tab:d24_preliminary_short}
\end{table}

\end{landscape}

\section{Full LM benchmark results for other model sizes}

\label{app:d12}
\begin{landscape}
\subsection{Depth-12 Models}
\begin{center}
\centering
\tiny
\setlength{\tabcolsep}{2pt}
\resizebox{18cm}{!}{%
\begin{tabular}{lp{2.2em}p{2.2em}p{2.2em}p{2.2em}p{2.2em}p{2.2em}p{2.2em}p{2.2em}p{2.2em}p{2.2em}p{2.2em}p{2.2em}p{2.2em}p{2.2em}p{2.2em}p{2.2em}p{2.2em}p{2.2em}p{2.2em}p{2.2em}p{2.2em}p{2.2em}p{2.2em}p{2.2em}p{2.2em}p{2.2em}p{2.2em}}
\toprule
 & \multicolumn{6}{c}{\textbf{World Knowledge}} & \multicolumn{7}{c}{\textbf{Commonsense}} & \multicolumn{4}{c}{\textbf{Passage Comprehension}} & \multicolumn{5}{c}{\textbf{Symbolic reasoning}} &  &  &  &  &  \\
\cmidrule(lr){2-7}\cmidrule(lr){8-14}\cmidrule(lr){15-18}\cmidrule(lr){19-23}
\textbf{Method} & \rotatebox{60}{\textbf{jeopardy}} & \rotatebox{60}{\textbf{bb-qa-wikidata}} & \rotatebox{60}{\textbf{arc-easy}} & \rotatebox{60}{\textbf{arc-challenge}} & \rotatebox{60}{\textbf{openbook-qa}} & \rotatebox{60}{\textbf{bb-language-identification}} & \rotatebox{60}{\textbf{copa}} & \rotatebox{60}{\textbf{commonsense-qa}} & \rotatebox{60}{\textbf{piqa}} & \rotatebox{60}{\textbf{hellaswag}} & \rotatebox{60}{\textbf{hellaswag-zeroshot}} & \rotatebox{60}{\textbf{winograd}} & \rotatebox{60}{\textbf{winogrande}} & \rotatebox{60}{\textbf{squad}} & \rotatebox{60}{\textbf{coqa}} & \rotatebox{60}{\textbf{boolq}} & \rotatebox{60}{\textbf{lambada-openai}} & \rotatebox{60}{\textbf{bb-dyck-languages}} & \rotatebox{60}{\textbf{bb-cs-algorithms}} & \rotatebox{60}{\textbf{bb-repeat-copy-logic}} & \rotatebox{60}{\textbf{bb-operators}} & \rotatebox{60}{\textbf{lsat-ar}} & \rotatebox{60}{\textbf{CORE avg}} & \rotatebox{60}{\textbf{MMLU (ICL)}} & \rotatebox{60}{\textbf{GSM8K (p@5)}} & \rotatebox{60}{\textbf{GSM8K (strict)}} & \rotatebox{60}{\textbf{Lambada (PPL)}} \\
\midrule
\multicolumn{28}{l}{\textit{Baselines}} \\
\bottomrule
\end{tabular}
}
\captionsetup{font=footnotesize,width=22cm}
\captionof{table}{depth-12 results: full CORE benchmark suite (per-task ICL accuracy), CORE average, MMLU (ICL), GSM8K pass@5 under strict and relaxed matchers and Lambada perplexity and accuracy; mean $\pm$ std over three seeds for every row. Every MTP model uses $k{=}4$ targets. Every run is trained with parameter-matched budget, so the number of steps follows each model's parameter count. Meta-L therefore trains on 68.6\% and DeepSeek MTP on 22.5\% more tokens than NTP, and Meta on 19.3\%, the most likely explanation for their improvements over NTP at this size. CORE (centered) averages each of the 22 CORE tasks as $(\mathrm{acc}-b)/(1-b)$ with respect to random baseline $b$. \textbf{LAMBADA columns.} The base model (not a Lambada-finetuned variant) on the LAMBADA test split: target-word perplexity (lower is better) and accuracy.}
\label{tab:d12_preliminary_short}
\end{center}

\end{landscape}

\begin{landscape}
\subsection{Depth-18 Models}
\label{app:d18}
\begin{table}[!b]
\centering
\tiny
\setlength{\tabcolsep}{2pt}
\resizebox{18cm}{!}{%
\begin{tabular}{lp{2.2em}p{2.2em}p{2.2em}p{2.2em}p{2.2em}p{2.2em}p{2.2em}p{2.2em}p{2.2em}p{2.2em}p{2.2em}p{2.2em}p{2.2em}p{2.2em}p{2.2em}p{2.2em}p{2.2em}p{2.2em}p{2.2em}p{2.2em}p{2.2em}p{2.2em}p{2.2em}p{2.2em}p{2.2em}p{2.2em}p{2.2em}}
\toprule
 & \multicolumn{6}{c}{\textbf{World Knowledge}} & \multicolumn{7}{c}{\textbf{Commonsense}} & \multicolumn{4}{c}{\textbf{Passage Comprehension}} & \multicolumn{5}{c}{\textbf{Symbolic reasoning}} &  &  &  &  &  \\
\cmidrule(lr){2-7}\cmidrule(lr){8-14}\cmidrule(lr){15-18}\cmidrule(lr){19-23}
\textbf{Method} & \rotatebox{60}{\textbf{jeopardy}} & \rotatebox{60}{\textbf{bb-qa-wikidata}} & \rotatebox{60}{\textbf{arc-easy}} & \rotatebox{60}{\textbf{arc-challenge}} & \rotatebox{60}{\textbf{openbook-qa}} & \rotatebox{60}{\textbf{bb-language-identification}} & \rotatebox{60}{\textbf{copa}} & \rotatebox{60}{\textbf{commonsense-qa}} & \rotatebox{60}{\textbf{piqa}} & \rotatebox{60}{\textbf{hellaswag}} & \rotatebox{60}{\textbf{hellaswag-zeroshot}} & \rotatebox{60}{\textbf{winograd}} & \rotatebox{60}{\textbf{winogrande}} & \rotatebox{60}{\textbf{squad}} & \rotatebox{60}{\textbf{coqa}} & \rotatebox{60}{\textbf{boolq}} & \rotatebox{60}{\textbf{lambada-openai}} & \rotatebox{60}{\textbf{bb-dyck-languages}} & \rotatebox{60}{\textbf{bb-cs-algorithms}} & \rotatebox{60}{\textbf{bb-repeat-copy-logic}} & \rotatebox{60}{\textbf{bb-operators}} & \rotatebox{60}{\textbf{lsat-ar}} & \rotatebox{60}{\textbf{CORE avg}} & \rotatebox{60}{\textbf{MMLU (ICL)}} & \rotatebox{60}{\textbf{GSM8K (p@5)}} & \rotatebox{60}{\textbf{GSM8K (strict)}} & \rotatebox{60}{\textbf{Lambada (PPL)}} \\
\midrule
\multicolumn{28}{l}{\textit{Baselines}} \\
\bottomrule
\end{tabular}
}
\captionsetup{font=footnotesize}
\captionof{table}{depth-18 results: full CORE benchmark suite (per-task ICL accuracy), CORE average, MMLU (ICL), GSM8K pass@5 under both matchers and Lambada perplexity and accuracy; mean $\pm$ std over three seeds for every row. Every MTP model uses $k{=}4$ targets; ``LightMTP'' layer 12 is d24's target layer 16 scaled to depth 18, layer 2 is kept as is. Every run is trained with a parameter-matched budget, so the number of steps follows each model's parameter count; Meta MTP therefore trains on 34.9\% and DeepSeek MTP on 17.2\% more tokens than NTP. Cells color-coded vs NTP. CORE (centered) averages each of the 22 CORE tasks as $(\mathrm{acc}-b)/(1-b)$ with respect to random baseline $b$. \textbf{Lambada columns.} The base model (not a Lambada-finetuned variant) on the LAMBADA test split: target-word perplexity (lower is better) and accuracy.}
\label{tab:d18_preliminary_short}
\end{table}

\end{landscape}

\end{document}